\pdfoutput=1
\PassOptionsToPackage{table}{xcolor}
\documentclass{article} 
\usepackage[preprint]{colm2026_conference}

\usepackage{microtype}
\usepackage{amsmath,amssymb}
\usepackage{placeins}
\setcitestyle{numbers,square}
\usepackage{geometry}
\usepackage{hyperref}
\usepackage{url}
\usepackage{booktabs}
\usepackage{lineno}
\usepackage{graphicx}
\usepackage{tocloft}

\usepackage{makecell}
\usepackage{multirow}
\usepackage{colortbl}
\usepackage[table]{xcolor}
\usepackage{adjustbox}
\usepackage{array}
\newcolumntype{L}[1]{>{\raggedright\arraybackslash}p{#1}}

\definecolor{hunyuanblue}{RGB}{232,242,255}

\newcommand{\hybr}{\rowcolor{hunyuanblue}}

\newcommand{\orglogo}[1]{\raisebox{-0.2\height}{#1}\,}

\newcommand{\nextline}{\\} 

\usepackage{venusocr_report}

\title{\textcolor{venuslogo}{VenusOCR~2.0}: Towards Efficient and Effective Document Parsing}

\author{%
{\normalsize\bfseries Lei Li$^{\dagger}$, Zheng Wei, Yi Yuan, Qingbin Liu$^{\ddagger}$, Jiang Bian, Zang Li}\\[5pt]
{\small{VenusOCR Team, Tencent Inc.}}%
}

\setreportlinks{%
  \venuslinkpill{\includegraphics[height=9pt]{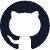}}{https://github.com/flow3rdown/VenusOCR}{\textbf{Code:} \textcolor{venusnavy}{github.com/flow3rdown/VenusOCR}}\\[4.5pt]%
  \venuslinkpill{\includegraphics[height=9pt]{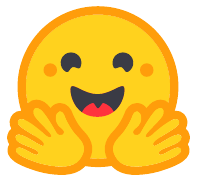}}{https://huggingface.co/flow3rdown/VenusOCR}{\textbf{Model:} \textcolor{venusnavy}{huggingface.co/flow3rdown/VenusOCR}}%
}

\hypersetup{pdftitle={VenusOCR 2.0 Technical Report},
            pdfauthor={VenusOCR Team}}

\begin{document}
\ifcolmsubmission
    \linenumbers
\fi
\maketitle
\thispagestyle{firststyle}

\begin{abstract}
    We present \textbf{VenusOCR~2.0}, a compact 1B-parameter vision-language recognizer designed for both high-accuracy and high-throughput document parsing across text, tables, and formulas.
To continuously improve recognition quality without self-training collapse, we build a bounded recursive self-improvement (RSI) data engine where the model acts strictly as a weakness probe,
while external multi-expert medoid consensus, VLM visual verification, and factorized synthesis supply reliable supervision across pretraining, curated SFT, and RLVR.
VenusOCR~2.0 couples a dynamic-resolution SigLIP2 encoder and $2\times2$ spatial token merging with a Qwen3-0.6B decoder, adopting compact atomic OTSL tokens to shorten table sequences
and reduce structural errors. To further eliminate the autoregressive decoding bottleneck, we introduce \textbf{PTPv2}, a lightweight ($\sim$67M) decoupled parallel
draft module that combines frozen multi-layer verifier conditioning, a standalone learned mask, and block-causal all-position next-token prediction.
On OmniDocBench~v1.6, VenusOCR~2.0 achieves an overall score of 96.45 and a state-of-the-art 98.57 formula CDM—substantially outperforming massive general-purpose VLMs such as Gemini~3~Pro (92.91)
and Qwen3-VL-235B (89.78). Deployed in vLLM on a single H20 GPU, PTPv2 delivers a $2.96\times$ single-request speedup and up to $2.36\times$ higher
batched throughput while guaranteeing lossless, token-for-token equivalence with standard autoregressive decoding.
\end{abstract}
\begingroup
\renewcommand{\thefootnote}{\fnsymbol{footnote}}%
\footnotetext[2]{Project leader.}%
\footnotetext[3]{Corresponding author.}%
\endgroup

\vspace{-4pt}
\begin{center}
\begin{minipage}{\linewidth}
    \centering
    \includegraphics[width=0.90\linewidth]{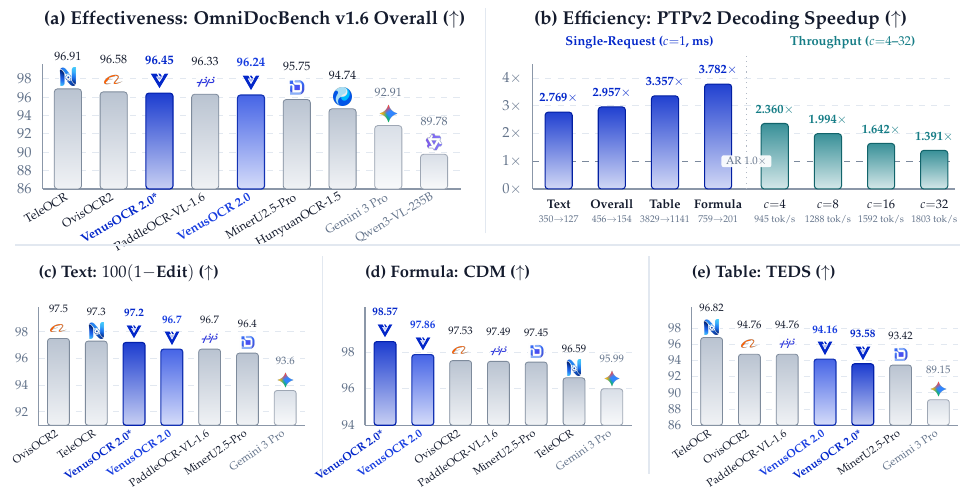}
    \captionof{figure}{\textbf{Overview of VenusOCR~2.0.} (a)~OmniDocBench~v1.6 overall accuracy against specialized OCR models
        and general VLMs (default: PP-DocLayoutV3; $*$: paired with Tele-layout). (b)~Lossless PTPv2 decoding speedups over autoregressive
        serving on a single H20 GPU across single-request latency and batched throughput. (c--e)~Category scores on
        text, formula, and table recognition.}
    \label{fig:hero_overview}
\end{minipage}
\end{center}

\clearpage
\setcounter{tocdepth}{2}
\tableofcontents
\clearpage

\section{Introduction}
\label{sec:introduction}

Document parsing converts PDFs, scans, and camera-captured pages into structured formats for
retrieval-augmented generation, automated
workflows, and agentic tasks~\citep{MinerU_2024,OmniDocBench_2025,olmOCR_2025,OCRHindersRAG_2024,DocParsing_Survey_2024}.
Unlike plain text OCR, faithful parsing requires accuracy across complex layout structures, nested
mathematical expressions, and spanning table topology, where dropping a single minus sign, brace,
or column boundary breaks downstream semantics. Vision-language models (VLMs) have largely replaced
brittle rule-based OCR pipelines, evolving along both layout-decoupled
architectures~\citep{MinerU2_5_2025,MonkeyOCR_2025,Dolphin_2025,PaddleOCR_VL_1_6_2026} and
end-to-end formulations~\citep{GOT_OCR2_2024,HunyuanOCR_2025,dots_ocr_2025,GLM_OCR_2026}. Yet
building a compact ($\sim$1B) parser that matches frontier general VLMs while serving at high throughput
remains challenging in terms of both usability and practicality.

First, naive data scaling or self-training quickly hits diminishing returns in compact models.
Parsing errors typically come from missing coverage of rare layouts and degradations, noisy labels
in public datasets or pseudo-labels, and complex structural targets that token-level
cross-entropy only weakly captures. Recent curation pipelines use multi-model
agreement~\citep{ConsensusEntropy_2025,MinerU2_5_Pro_2026}, targeted region
refinement~\citep{PaddleOCR_VL_1_6_2026,Xiaomi_OCR_0_2026}, and agent-assisted data
flows~\citep{HunyuanOCR_1_5_2026}. However, treating every high-loss or model-disagreed sample as a
useful hard example conflates label noise with model weakness, and naive augmentation often
duplicates existing failure patterns instead of varying the underlying structural or visual
factors. Second, a compact backbone alone does not eliminate autoregressive (AR) decoding latency.
High-resolution crops expand visual token counts, and complex tables or multi-line formulas
generate long structured outputs one token at a
time~\citep{SpecDec_Leviathan_2023,PTP_2026,Unlimited_OCR_2026}. 
To improve the throughput of document parsing models, we previously proposed the Parallel Token Prediction (PTP) method~\citep{PTP_2026}.
However, it still executes proposal steps through the full target backbone, incurring non-negligible latency.
Recently, decoupled block-diffusion drafting methods, such as DFlash~\citep{DFlash_2026,HunyuanOCR_1_5_2026}, 
have offloaded drafting to a shallow side network. 
While this significantly accelerates speculative decoding, it remains ill-suited for direct application 
in document parsing. This is because its anchor-skipping layout reserves an unused slot, 
reuses frozen vocabulary embeddings as masks, and relies on bidirectional block attention.

To address these issues, we release \textbf{VenusOCR~2.0}, a 1B-parameter vision-language
recognizer for text, tables, and formulas within a layout-guided parsing pipeline. On the data side
(Section~\ref{sec:data}), we build a bounded recursive self-improvement (RSI) data engine in which
the current recognizer acts strictly as a weakness probe rather than a label
source. External 4--7 expert medoid consensus under task-matched metrics
(NED~\citep{Levenshtein_1966}, TEDS~\citep{PubTabNet_TEDS_2020}, CDM~\citep{CDM_2024}), 27B VLM
visual verification, and factorized
$\mathrm{Content}\times\mathrm{Style}\times\mathrm{Degradation}$ synthesis provide verified targets across modality alignment, pretraining, curated SFT, and
metric-aware RL. On the model side (Section~\ref{sec:model}), our VenusOCR~2.0 model pairs a dynamic-resolution SigLIP2-style
encoder~\citep{NaViT_2023,SigLIP2_2025} and a $2\times2$ spatial merger with a Qwen3-0.6B
decoder~\citep{Qwen3_2025} using M-RoPE and atomic OTSL table serialization~\citep{OTSL_2023}.
After SFT, Stage~4 optimizes task metrics directly via RL with
soft penalties for malformed outputs. For fast inference without modifying verifier weights or
output distributions, we introduce \textbf{PTPv2}: a shallow 4-layer ($\sim$67M) draft conditioned
on frozen multi-layer verifier features that replaces anchor-skipping vocabulary masks with a
standalone learned mask vector $e_{\mathrm{mask}}$, applies fixed-anchor block-causal attention
($W=2048$), and supervises every query slot under D-PACE prefix-weighted loss ($B$ queries for $B$
proposals).

On OmniDocBench~v1.6~\citep{OmniDocBench_2025,MinerU2_5_Pro_2026}, VenusOCR~2.0 improves steadily
from 94.96 overall after pretraining (Stage~2) to 95.69 after curated SFT and 96.24 after metric RL.
When paired with Tele-layout, 
it achieves an overall score of 96.45 and a formula CDM of 98.57, substantially outperforming much 
larger general VLMs—such as Gemini~3~Pro (92.91)~\citep{Gemini_3_2026} 
and Qwen3-VL-235B (89.78)~\citep{Qwen3_VL_2025}—while matching top-tier specialized parsers.
In terms of efficiency, our 5-proposal PTPv2 draft
attains 91.91\% offline teacher agreement, delivers a $2.96\times$ single-request latency speedup and
a $2.95\times$ wall-clock throughput speedup ($3.78\times$ on formulas and $3.36\times$ on tables)
along with $1.39$--$2.36\times$ throughput gains across concurrency 4--32, and preserves 100\%
token-for-token equivalence with AR decoding on all non-truncated requests.

Our main contributions are as follows:
\begin{itemize}
    \item \textbf{Bounded RSI Data Engine:} We separate weakness probing ($\mathcal{M}_t$) from label
          generation via 4--7 expert medoid consensus, VLM render verification, and
          $\mathrm{Content}\times\mathrm{Style}\times\mathrm{Degradation}$ synthesis across pretraining, SFT,
          and metric RL.
    \item \textbf{Compact 1B Recognizer:} We design a lightweight vision-language architecture coupling a dynamic-resolution SigLIP2 and Qwen3-0.6B backbone with
          $2\times2$ spatial token merging, M-RoPE, and atomic OTSL table serialization. Through stage-by-stage training, 
          it outperforms massive general VLMs (e.g., 235B) while rivaling leading specialized parsers.
    \item \textbf{PTPv2 Parallel Drafting:} We introduce a $\sim$67M decoupled draft combining multi-layer
          verifier feature conditioning, a standalone learned mask, fixed-anchor block-causal attention, and
          all-position next-token supervision with exact AR equivalence.
\end{itemize}

\section{Related Work}
\label{sec:related}

\paragraph{Document Parsing.}
Document parsing has evolved from traditional multi-stage OCR pipelines and early OCR-free
transformers~\citep{Donut_2022,Nougat_2023,DocParsing_Survey_2024} to specialized
VLMs~\citep{GOT_OCR2_2024,OCR_FM_Survey_2026}, broadly categorized into layout-decoupled
systems~\citep{DocLayout_YOLO_2024,MinerU_2024,MinerU2_5_2025,MinerU2_5_Pro_2026,MonkeyOCR_2025,Dolphin_2025,PaddleOCR_VL_2025,PaddleOCR_VL_1_6_2026}
and unified end-to-end
models~\citep{olmOCR_2025,Mistral_OCR_2025,SmolDocling_2025,HunyuanOCR_2025,dots_ocr_2025,DeepSeek_OCR_2025,DeepSeek_OCR_2_2026,OCRVerse_2026}.
Across both paradigms, compact 0.8B--4B
specialists~\citep{GLM_OCR_2026,OvisOCR2_2026,TeleOCR_2026,FireRed_OCR_2026,Logics_Parsing_2025}
frequently outperform much larger general
VLMs~\citep{Qwen3_VL_2025,Gemini_3_2026,Kimi_K2_5_2026,Ovis2_5_2025,InternVL3_5_2025}.
Achieving such high accuracy in compact document VLMs depends heavily on corpus diversity, label
quality~\citep{MinerU2_5_Pro_2026,PaddleOCR_VL_1_6_2026,HunyuanOCR_1_5_2026,Xiaomi_OCR_0_2026}, and advanced post-training.
Recent data pipelines combine multi-model
agreement~\citep{ConsensusEntropy_2025,MinerU2_5_Pro_2026,Xiaomi_OCR_0_2026}, automated VLM
diagnostics~\citep{DOCR_Inspector_2025}, agentic data workflows~\citep{HunyuanOCR_1_5_2026}, and
realistic degradation synthesis~\citep{Augraphy_2023,RealisticSceneSynthesis_2026}. On the
optimization side, verifiable reinforcement learning~\citep{GRPO_2024,DAPO_2025} with structural
and format-decoupled
rewards~\citep{olmOCR_2025,olmOCR_2_2025,Infinity_Parser_2025,FormatDecoupledRL_2026,MinerU2_5_Pro_2026,PaddleOCR_VL_1_6_2026}
closes structural gaps remaining after SFT on OCR
benchmarks~\citep{OCRBench_v2_2025,OmniDocBench_2025}. VenusOCR~2.0 builds on these ideas by
separating label uncertainty from model difficulty via task-matched medoid consensus, targeting
structural gaps with factorized synthesis, and directly optimizing NED, TEDS, and CDM in Stage~4
RL.

\paragraph{Efficient Inference and Parallel Decoding.}
Despite the compactness of specialized document VLMs, parsing dense pages still incurs high latency due to lengthy visual inputs and autoregressive (AR) generation. Recent
methods lower per-step cost through visual token and KV-cache
compression~\citep{NaViT_2023,SigLIP2_2025,DeepSeek_OCR_2025,PaddleOCR_VL_C2F_2026,FastOCR_2026,Unlimited_OCR_2026}
or compact representations like atomic table serialization~\citep{OTSL_2023}. Orthogonal to token compression, speculative and
parallel decoding~\citep{SpecDec_Leviathan_2023,Gloeckle_MTP_2024,BlockDiffusion_2025}---including
OCR multi-token prediction, hierarchical drafting, diffusion OCR, and
DFlash~\citep{PTP_2026,P_MTP_2026,HSD_2026,GravityOCR_2026,DFlash_2026,HunyuanOCR_1_5_2026}---cuts
sequential AR steps directly. PTPv2 bridges PTP's all-position next-token contract~\citep{PTP_2026}
with DFlash's shallow feature-conditioned draft~\citep{DFlash_2026}, eliminating anchor-skipping
slots for clean deployment in vLLM~\citep{vLLM_2023,FlashAttention2_2023}.

\section{Recursive Self-Improvement Data Engine}
\label{sec:data}

\begin{figure}[t]
    \centering
    \includegraphics[width=0.98\linewidth]{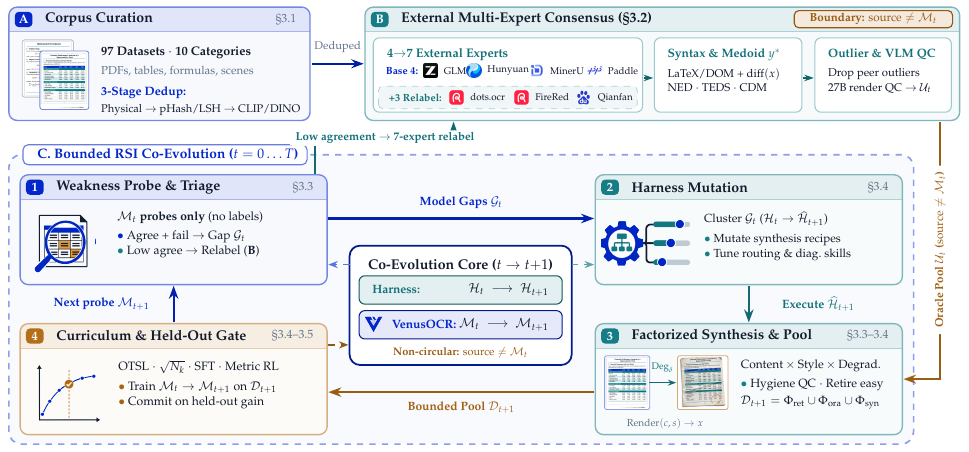}
    \caption{Bounded recursive self-improvement (RSI) data engine. Corpus curation~(A) and multi-expert
        consensus~(B) supply non-circular supervision ($\mathrm{source}\neq\mathcal{M}_t$), while the four-step
        RSI cycle~(C) iteratively probes model weaknesses, updates the data harness $\mathcal{H}_t$, synthesizes
        and filters training pools, and gates model commits on held-out validation.}
    \label{fig:dataengine}
\end{figure}

Scaling a compact (1B) document recognizer by simply adding more data quickly yields diminishing
returns and amplifies label noise, while naive self-training reinforces a student model's own
mistakes. VenusOCR~2.0 introduces a bounded recursive self-improvement (RSI) data
engine (Figure~\ref{fig:dataengine}). Across training stages, the current model
$\mathcal{M}_t$ serves strictly as a diagnostic probe to expose residual errors, decoding collapses, and
coverage gaps---never as a direct source of pseudo-labels. Driven by these diagnostics, the engine
co-evolves both the recognizer weights and its own data-production harness (synthesis recipes,
diagnostic playbooks, routing policies, and curriculum mixtures), while grounding every target in
external multi-expert consensus, VLM visual verification, and pre-degradation synthesis. Specifically,
our pipeline integrates corpus curation (\S\ref{sec:curation}), consensus verification
(\S\ref{sec:consensus}), targeted weakness mining (\S\ref{sec:diagnosis}), closed-loop
orchestration (\S\ref{sec:rsi_loop}), and a stage-aligned curriculum (\S\ref{sec:curriculum}).

\subsection{Corpus Curation and Deduplication}
\label{sec:curation}

\begin{figure}[t]
    \centering
    \includegraphics[width=0.98\linewidth]{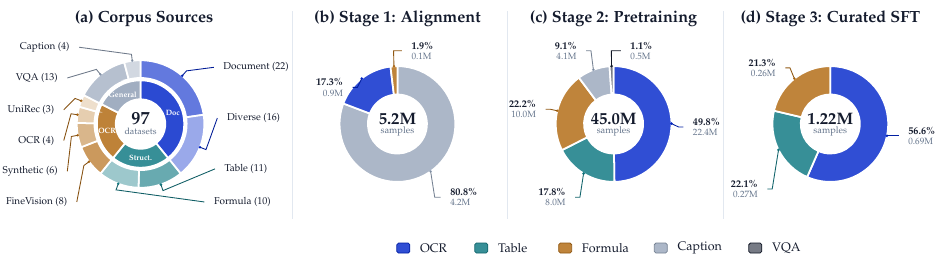}
    \caption{Training corpus composition and curriculum mixture. (a)~Taxonomy of pretraining data sources
        grouped into four functional pillars and ten categories. (b--d)~Task-level sample distribution across the
        alignment, pretraining, and curated SFT stages.}
    \label{fig:data_distribution}
\end{figure}

To prevent iterative mining from cycling over repeated document templates, we consolidate 97
sub-datasets across ten source categories into a unified schema. The corpus covers real
PDFs~\citep{olmOCR_2025,FinePDFs_2025,DocLayNet_2022,CDLA_2021},
cropped tables~\citep{PubTables1M_2022,FinTabNet_GTE_2021,SciTSR_2019},
formulas~\citep{UniMERNet_2024,MathWriting_2025}, printed and handwritten text
OCR~\citep{CASIA_HWDB_2011,GNHK_2021,DocBank_2020}, general vision-language
pairs~\citep{FineVision_2025,OpenBee_2025,LAION5B_2022,Wukong_2022,DataComp_2023}, diverse scenes
such as receipts, slides, posters, screenshots, and newspapers~\citep{CORD_2019}, and targeted
synthetic data (Figure~\ref{fig:data_distribution}a).
We apply a three-stage deduplication and filtering pipeline: (1)~physical filtering removes
corrupted PDFs and crops with extreme aspect ratios; (2)~perceptual hashing
(intra-document pHash, multi-table LSH pHash, and MinHash LSH over full PDFs, crops, and
layout-detected blocks~\citep{PP_DocLayout_2025}) remove or downsample repeated templates; and
(3)~cluster-based filtering, which leverages CLIP and DINO embeddings~\citep{CLIP_2021,DINOv2_2023} indexed via FAISS IVF-Flat, 
applies hierarchical $k$-means farthest-point sampling to prune semantic near-duplicates.

\subsection{Task-Aware Consensus and Verification}
\label{sec:consensus}

To address the lack of ground truth in unlabeled documents and the persistent noise in public datasets,
we distill reliable supervision from an ensemble of external OCR experts.
We start with a base pool of four experts (GLM, Hunyuan, MinerU, 
and Paddle~\citep{GLM_OCR_2026,HunyuanOCR_2025,MinerU2_5_2025,PaddleOCR_VL_2025}) 
to establish initial consensus labels, and expand to seven (incorporating dots.mocr, 
FireRed-OCR, and Qianfan~\citep{dots_mocr_2026,FireRed_OCR_2026,Qianfan_OCR_2026}) 
for targeted relabeling of ambiguous cases~\citep{ConsensusEntropy_2025}.
Before aggregation, candidate outputs undergo strict syntax validation—such as LaTeX tokenization 
for formulas and DOM parsing for tables—to filter out malformed generations, severe repetitions, 
and execution timeouts.
Given a set of valid candidates $\mathcal{V}(x)$ ($|\mathcal{V}|\ge 2$), we identify the consensus $y^*$ 
as the medoid under a
task-matched disagreement metric $d_t$~\citep{Levenshtein_1966,PubTabNet_TEDS_2020,CDM_2024}:
\begin{align}
    y^*      & = y_{\arg\min_{i\in\mathcal{V}} D_i}, \qquad
    D_i = \frac{1}{|\mathcal{V}|-1}\sum_{j\in\mathcal{V},\,j\ne i} d_t(y_i, y_j), \label{eq:medoid} \\
    d_t(a,b) & =
    \begin{cases}
        \operatorname{NED}(a,b),    & t=\mathrm{text},    \\
        1-\operatorname{TEDS}(a,b), & t=\mathrm{table},   \\
        1-\operatorname{CDM}(a,b),  & t=\mathrm{formula},
    \end{cases}
    \;\;\;
    \mathrm{diff}(x) = 0.5(1{-}c) + 0.3\,\delta_{\max} + 0.2\,r_{\mathrm{inv}},
    \label{eq:taskdistance}
\end{align}
where $c=1-D_{y^*}$, $\delta_{\max}=\max_{i,j\in\mathcal{V}} d_t(y_i,y_j)$, and $r_{\mathrm{inv}}$
is the weighted invalid ratio. 
To stratify data quality, we compute a composite difficulty score $\mathrm{diff}(x)$, which integrates 
the overall consensus level ($1-D_{y^*}$), the maximum pairwise disagreement among experts, 
and the proportion of invalid outputs. This metric allows us to partition the dataset into easy, 
medium, hard, and noisy tiers. For the expanded 7-expert pool, we apply an outlier pruning step: 
experts that significantly underperform the peer consensus are dropped before recomputing the 
difficulty score, which helps recover valid hard samples from false noise.
Subsequently, an automated visual verification stage renders the candidate texts, formulas, 
and tables alongside the source image crops, utilizing a specialized multimodal 
judge~\citep{DOCR_Inspector_2025} to distinguish between genuine annotation errors and degraded inputs. 
Finally, for extreme edge cases that exhibit severe expert disagreement, we 
employ a general-purpose VLM as a fallback arbiter to either re-annotate or discard the samples. 
In practice, we find that such low-consensus instances primarily stem from fundamental defects in 
the source images, such as severe blur or incomplete cropping.

\subsection{Weakness Mining and Factorized Synthesis}
\label{sec:diagnosis}\label{sec:agent}

To steer data collection toward genuine model weaknesses, the target model $\mathcal{M}_t$ 
periodically performs inference across both real and synthetic data pools. 
We identify systematic blind spots by jointly monitoring prediction residuals, dataset-level loss trends, 
generation degeneracies (e.g., severe repetition), and reinforcement learning reward signals (Figure~\ref{fig:dataengine}, bottom).

\paragraph{Reliability-conditioned routing.}
Rather than treating all high-loss examples equally, we cross-reference $\mathcal{M}_t$'s failures
against the multi-expert consensus established earlier. 
A sample is routed directly for model refinement only if $\mathcal{M}_t$ fails despite strong agreement among the external experts,
split between valid low-similarity cases (to improve accuracy) and repetition or syntax collapses (to stabilize output
structure). Conversely, when a sample lacks a reliable external consensus, 
it is diverted back to our annotation pipeline for expanded relabeling, visual inspection, 
or outright exclusion.

\paragraph{Agent-assisted diagnosis and factorized synthesis.}
For systematic coverage gaps such as dense spanning tables, nested formulas, or degraded scans, an
agent-assisted workflow turns failure diagnostics into synthesis
specification.  Specifically,  the agent extracts structural descriptors (e.g., cell span patterns, inline math density, and formatting conventions) 
from failure seeds to guide syntax-checked LLM synthesis. Generated content
feeds into a factorized rendering pipeline that decouples semantic content $c$, visual style $s$, and degradation
$\delta$~\citep{Augraphy_2023,RealisticSceneSynthesis_2026}:
\begin{equation}
    x = \operatorname{Degrade}_{\delta}\!\left(\operatorname{Render}(c, s)\right), \qquad y = \operatorname{Serialize}_t(c).
    \label{eq:synthesis}
\end{equation}
For table synthesis, we pair real structural skeletons with diverse cell contents and visual themes. 
A headless browser engine renders these into clean images strictly aligned with the ground-truth targets $y$,
before applying realistic visual degradations under strict legibility and layout constraints. 
As this synthesis pipeline focuses on localized, element-level crops, broader page-level layout understanding 
and reading order are learned from real documents paired with external layout detectors.

\subsection{Closed-Loop Orchestration: Automating the Bounded RSI Cycle}
\label{sec:rsi_loop}

While Sections~\ref{sec:curation}--\ref{sec:diagnosis} describe the individual modules for
deduplication, external consensus, and factorized synthesis, a static one-off pipeline only yields
a single improvement pass. True recursive self-improvement (RSI) requires not only that the
recognizer $\mathcal{M}_t$ improves, but that the system's capability to improve itself---its
data-production harness $\mathcal{H}_t$---also becomes sharper across cycles. To achieve
this without sacrificing label integrity, VenusOCR~2.0 orchestrates Sections~\ref{sec:curation}--\ref{sec:diagnosis}
as a \emph{co-evolving harness-and-model} loop across iterations $t = 0, 1, \dots, T$
(Figure~\ref{fig:dataengine}). Concretely, each cycle jointly upgrades the data-engine harness
$\mathcal{H}_t$ (how failures are diagnosed, routed, and synthesized) and the recognizer weights
$\mathcal{M}_t$ (how documents are parsed), while keeping recursion strictly \emph{bounded} by an
independent verification boundary: neither $\mathcal{M}_t$ nor the harness-editing agent grades its
own outputs, and all targets come solely from external multi-expert consensus,
render-based VLM checks, or programmatic ground truth ($\mathrm{source} \neq \mathcal{M}_t$).

Each automated iteration runs a continuous explore--propose--verify--commit cycle in four steps:
\begin{enumerate}
    \item \textbf{Failure exploration and triage.} The orchestrator runs $\mathcal{M}_t$ over
          designated probing slices and curated corpus shards, then
          cross-references prediction residuals against external consensus confidence.
          This separates genuine model capability gaps (high expert agreement, weak
          $\mathcal{M}_t$ prediction) from annotation ambiguity (high expert disagreement),
          sending ambiguous items back to expanded 7-expert relabeling and VLM inspection rather
          than mistaking label noise for hard examples.
    \item \textbf{Harness mutation and skill accumulation ($\mathcal{H}_t \rightarrow \widehat{\mathcal{H}}_{t+1}$).}
          Rather than merely re-weighting a fixed dataset, a diagnostic agent clusters confirmed
          model-gap cases by failure mode, compares them against corpus statistics,
          and directly upgrades the data-engine harness $\mathcal{H}_t$.
          Specifically, it mutates synthesis recipes and configs (enriching table
          skeletons, expanding rare LaTeX macro grammars, and adjusting blur, skew, and lighting
          degradation profiles in Equation~\ref{eq:synthesis}), refines consensus and routing
          policies (reallocating 7-expert relabeling budget and adjusting domain-specific outlier
          filtering), and distills recurring failure patterns into reusable diagnostic templates so
          subsequent rounds identify subtle blind spots faster.
    \item \textbf{Independent verification and bounded pool assembly ($\mathcal{D}_{t+1}$).}
          Executing the candidate harness $\widehat{\mathcal{H}}_{t+1}$ produces newly mined and
          synthesized shards, all of which must pass independent hygiene gates---perceptual and
          semantic deduplication, syntax/DOM checks, and render
          verification---while samples already mastered by $\mathcal{M}_t$ are retired:
          \begin{equation}
              \mathcal{D}_{t+1} =
              \Phi_{\mathrm{retain}}(\mathcal{D}_t, \mathcal{M}_t)
              \;\cup\;
              \Phi_{\mathrm{oracle}}(\mathcal{U}_t, \mathcal{M}_t;\,\widehat{\mathcal{H}}_{t+1})
              \;\cup\;
              \Phi_{\mathrm{synth}}(\mathcal{G}_t;\,\widehat{\mathcal{H}}_{t+1}),
              \label{eq:rsi_pool}
          \end{equation}
          where $\Phi_{\mathrm{retain}}$ preserves non-saturated anchor data, $\Phi_{\mathrm{oracle}}$
          collects expert-verified gaps from candidate pool $\mathcal{U}_t$, and $\Phi_{\mathrm{synth}}$
          generates factorized synthetic data from diagnosed gap specifications $\mathcal{G}_t$.
    \item \textbf{Held-out evaluation and co-evolutionary commit ($(\mathcal{H}_t,\mathcal{M}_t)\rightarrow(\mathcal{H}_{t+1},\mathcal{M}_{t+1})$).}
          Training $\mathcal{M}_t$ on $\mathcal{D}_{t+1}$ distills the harness gains directly into
          the updated recognizer weights $\mathcal{M}_{t+1}$. To prevent the harness from
          overfitting to the probing split, $\mathcal{M}_{t+1}$ is evaluated on separate held-out
          validation suites across text, formula, and table slices; only harness edits that produce
          verified net gains without regression are committed as $\mathcal{H}_{t+1}$, after which
          $\mathcal{M}_{t+1}$ takes over as the probe for the next recursive cycle.
\end{enumerate}

\subsection{Data Normalization and Stage Curriculum}
\label{sec:curriculum}

\begin{table}[t]
    \centering
    \footnotesize
    \setlength{\tabcolsep}{3.5pt}
    \caption{Stage-aligned data curriculum (see Figure~\ref{fig:data_distribution}c--d for task-level breakdown).}
    \label{tab:data_curriculum}
    \begin{tabular}{@{}L{0.18\linewidth}L{0.27\linewidth}rL{0.31\linewidth}@{}}
        \toprule
        Stage             & Supervision focus            & Samples & Key shift                                      \\
        \midrule
        1: Alignment      & Captions + basic OCR/formula & 5.2M    & Projector-only modality alignment              \\
        2: Pretraining    & Multi-task + synthesis       & 45.0M   & Comprehensive and extensive \\
        3: Curated SFT    & Consensus + RSI synthesis    & 1.2M    & Hard samples                 \\
        4: Metric RL      & Non-saturated crops          & 20k     & Task-metric rewards $1-d_t$                    \\
        \bottomrule
    \end{tabular}
\end{table}

Before training, all mined, relabeled, and synthetic data streams undergo a unified formatting and hygiene
pipeline. Specifically, we transpile table structures into compact OTSL tokens~\citep{OTSL_2023}, 
and standardize mathematical LaTeX delimiters.
A rigorous filtering pass then strips residual HTML artifacts, 
control characters, and degenerate repetitive patterns. To improve robustness against orientation variances 
in real-world scans, we also apply stochastic orthogonal rotations to a small fraction of table crops during 
data loading.

As summarized in Table~\ref{tab:data_curriculum}, each training stage instantiates a progressively
sharper pass of the Bounded RSI loop.
Stage~1 warms up the modality projector using foundational image-caption and basic OCR pairs,
and Stage~2 scales up to full-model pretraining across the deduplicated multi-task corpus with
sub-linear mixture weighting.
Building on the Stage~2 checkpoint, Stage~3 executes a targeted Bounded RSI cycle that focuses
supervision on consensus-verified and factorized-synthetic hard examples, doubling the average
target sequence length. Finally, Stage~4 retires saturated samples relative
to the Stage~3 recognizer and applies metric-aware RL to the remaining non-saturated, challenging
instances.
Dealing with these data characteristics directly shaped our model architecture in Section~\ref{sec:model}. Specifically, we introduce dynamic 
visual merging to handle extreme aspect ratios, while adopting an atomic OTSL vocabulary and 
PTPv2 parallel decoding to efficiently generate long, token-heavy table sequences.

\section{Architecture, Training, and Parallel Decoding}
\label{sec:model}
\label{sec:system}

To handle wide crop aspect ratios and token-heavy tables and formulas, VenusOCR~2.0 pairs a compact
1B recognition backbone (\S\ref{sec:backbone}) with progressive post-training and metric-aware RL
(\S\ref{sec:training}), and accelerates inference via decoupled parallel drafting (PTPv2,
\S\ref{sec:ptpv2}).

\subsection{Architecture}
\label{sec:backbone}

VenusOCR~2.0 adopts a streamlined vision--projector--language architecture with approximately 1B parameters. 
To handle diverse document aspect ratios without distortion, we employ a dynamic-resolution SigLIP2 visual 
encoder~\citep{SigLIP2_2025,NaViT_2023} that processes high-resolution inputs at their native aspect ratios. 
Before feeding into the language backbone, a $2\times2$ spatial merger and MLP projector compress the visual sequence
by $4\times$. The resulting tokens are then processed by a Qwen3-0.6B decoder~\citep{Qwen3_2025}, 
which leverages multimodal rotary position embeddings (M-RoPE) to jointly capture 2D spatial layouts and 1D sequence order.

\begin{figure}[t]
    \centering
    \includegraphics[width=0.98\linewidth]{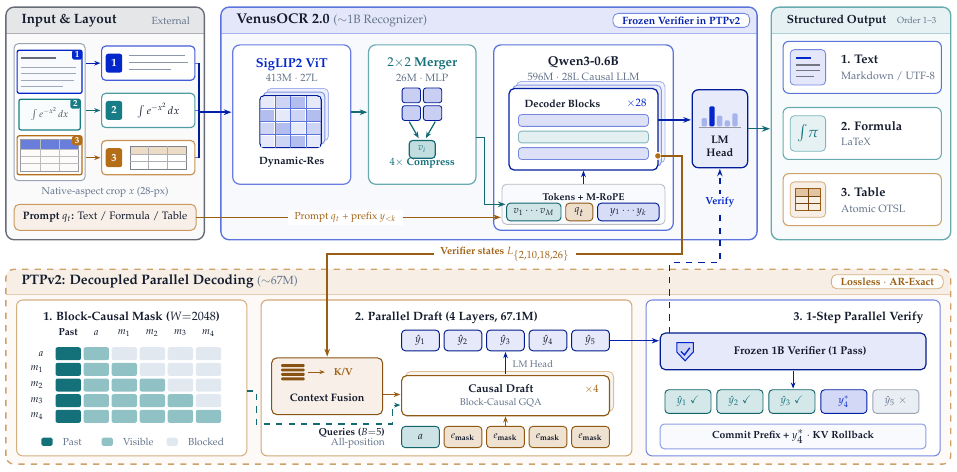}
    \caption{VenusOCR~2.0 architecture overview. An external layout detector supplies region crops and reading
        order to the 1B recognizer, while the optional PTPv2 draft module conditions on frozen verifier states
        for parallel decoding.}
    \label{fig:architecture}
\end{figure}

Following layout-decoupled parsing
paradigms~\citep{MinerU2_5_2025,PaddleOCR_VL_2025,DocLayout_YOLO_2024}, an external layout detector
supplies layout analysis, while the shared recognizer models the conditional distribution
$p_\theta(y\mid x,q_t)$
for image crop $x$ and task prompt $q_t$ (\texttt{Text Recognition:}, \texttt{Formula Recognition:}, or
\texttt{Table Recognition:}). Plain text and mathematical formulas are transcribed as standard
text and LaTeX, while tables are serialized into seven atomic Optimized Table Structure Language
(OTSL) tokens~\citep{OTSL_2023}. Replacing verbose HTML tags with atomic OTSL tokens substantially shortens sequences
and mitigates structural syntax errors, which is particularly beneficial for a compact language decoder. 
Keeping layout detection external separates segmentation
errors from recognition errors, leaving sequential token generation as the main serving bottleneck.
\subsection{Progressive Training}
\label{sec:training}

Following the staged data curriculum in Section~\ref{sec:curriculum}, VenusOCR~2.0 adapts the
pretrained SigLIP2 and Qwen3-0.6B backbones~\citep{SigLIP2_2025,Qwen3_2025} through four
progressive stages---from modality alignment to broad pretraining, curated SFT, and RLVR
(Reinforcement Learning with Verifiable Rewards). Across these stages, the trainable scope expands
from the projector alone to the full 1B recognizer, while supervision shifts from broad coverage
to consensus-verified hard cases and metric-aligned optimization.

\begin{table}[t]
    \centering
    \footnotesize
    \setlength{\tabcolsep}{4pt}
    \caption{Training stages across the VenusOCR~2.0 pipeline. Stages~2--4 update the 1B recognizer; PTPv2 is
        trained separately on frozen Stage~4 verifier features.} \label{tab:curriculum}
    \begin{tabular}{@{}L{0.20\linewidth}L{0.24\linewidth}L{0.48\linewidth}@{}}
        \toprule
        Stage                & Trainable scope           & Primary optimization objective                                 \\
        \midrule
        1: Alignment         & Projector only ($\sim$26M) & Align dynamic visual features with frozen LLM embeddings      \\
        2: Pretraining & Full recognizer (1B)      & Multi-task pretraining          \\
        3: Curated SFT       & Full recognizer (1B)      & Low-LR refinement on consensus hard cases and synthesis        \\
        4: Metric RL         & Full recognizer (1B)      & DAPO/GRPO optimization directly on task-matched metrics        \\
        \midrule
        PTPv2 (separate)     & Draft module ($\sim$67M)  & D-PACE weighted teacher-argmax CE on frozen verifier states    \\
        \bottomrule
    \end{tabular}
\end{table}

\paragraph{Stage 1: Modality Alignment.}
This stage freezes both the SigLIP2 visual encoder and the Qwen3-0.6B decoder, optimizing only the
randomly initialized vision--language projector. Training on foundational image--caption pairs mixed
with basic OCR and formula examples maps dynamic-resolution visual features into the language
embedding space and helps the projector adapt early to dense document typography before full-model
updates.

\paragraph{Stage 2: Multi-Task Pretraining and Context Extension.}
Stage~2 unfreezes the entire 1B recognizer using differential component learning rates---applying a
higher rate to the language decoder than to the vision encoder and projector---to adapt sequence
generation while preserving pretrained visual representations. Training covers the deduplicated
multi-task corpus across printed and handwritten text, OTSL-serialized tables, LaTeX formulas, and
retained general vision--language pairs using square-root mixture weighting.

\paragraph{Stage 3: Curated SFT.}
Rather than scaling data volume further, Stage~3 shifts to supervision quality and targeted weakness
repair, updating the full recognizer with a conservative, uniform learning rate. Guided by the
bounded RSI diagnostics (Section~\ref{sec:rsi_loop}), training concentrates on consensus-verified
hard cases, expanded multi-expert relabeling, and factorized synthetic crops covering dense
spanning tables, nested formulas, and degraded scans to resolve residual formatting, truncation,
and repetition errors.

\paragraph{Stage 4: Metric-Aligned RLVR.}
\label{sec:rl}
Token-level cross-entropy weights all token mismatches equally and poorly reflects global structure
such as table span topology or rendered formula equivalence. In Stage~4, we apply verifiable-reward
policy optimization~\citep{GRPO_2024,DAPO_2025,olmOCR_2_2025,Infinity_Parser_2025,FormatDecoupledRL_2026}
directly against the task-matched metrics corresponding to Equation~\eqref{eq:taskdistance}:
normalized edit similarity for text, TEDS (after OTSL-to-HTML conversion) for tables, and CDM for
formulas. Following DAPO~\citep{DAPO_2025}, optimization uses group-relative rollouts with
asymmetric clipping, token-level loss normalization, overlong-completion masking, and no KL penalty.
Standard OCR metrics often suffer from sharp reward drops when malformed outputs fail to parse or render, 
which can severely destabilize group-relative advantage estimation.
To mitigate this instability 
without compromising structural rigor, we assign a small positive floor reward to prediction-side syntax or 
rendering failures, rather than a hard zero. Conversely, invalid ground-truth references or empty generations 
still receive zero reward, ensuring that uninformative sample groups are cleanly zeroed out. 
Furthermore, we introduce a multiplicative penalty to discourage repetitive generations and restrict fast 
string-matching shortcuts in formulas to near-exact matches; this ensures that critical single-symbol errors 
(e.g., an omitted minus sign) are still rigorously evaluated via visual CDM scoring. 
Finally, as low-entropy OCR samples quickly saturate during training, 
we leverage the bounded RSI engine introduced in Section~\ref{sec:data} to automatically filter for non-saturated, verifiable crops.
\subsection{Parallel Token Prediction with PTPv2}
\label{sec:ptpv2}
\label{sec:efficiency}

Although visual token merging and atomic OTSL shorten sequences, standard autoregressive (AR)
decoding remains memory-bandwidth-bound, running a full 1B forward pass for every output token on
long tables and formulas. Existing parallel-decoding
methods~\citep{SpecDec_Leviathan_2023,Gloeckle_MTP_2024,BlockDiffusion_2025} make awkward
trade-offs here. Our earlier coupled PTP (PTPv1) framework~\citep{PTP_2026} inserts learnable register
tokens into the target sequence so every slot predicts future tokens, which still passes proposal
computation through the full target backbone. Decoupled block-diffusion methods such as
DFlash~\citep{DFlash_2026} move drafting to a shallow feature-conditioned module, substantially
reducing proposal latency. Inspired by this, we introduce PTPv2, a decoupled parallel-drafting
framework that pairs shallow verifier-conditioned drafting with PTP's all-position next-token
alignment (Figure~\ref{fig:architecture}, bottom, and Table~\ref{tab:ptpv2_compare}).

\begin{table}[t]
    \centering
    \footnotesize
    \caption{Comparison of decoding mechanisms. PTPv2 combines shallow decoupled drafting with all-position
        next-token alignment, a standalone learned mask, and block-causal visibility.}
    \label{tab:ptpv2_compare}
    \begin{tabular}{@{}L{0.145\linewidth}L{0.175\linewidth}L{0.17\linewidth}L{0.175\linewidth}L{0.135\linewidth}L{0.09\linewidth}@{}}
        \toprule
        Method                     & Proposal model               & Queries / $B$ tok           & Mask representation    & Block mask            & Verifier        \\
        \midrule
        Standard AR                & Full model (1B)              & Sequential ($1$/step)       & ---                    & Causal                & Frozen          \\
        PTPv1~\citep{PTP_2026}     & Coupled full model           & $B$ (register tokens)       & Learned registers      & Group-causal          & Modified        \\
        DFlash~\citep{DFlash_2026} & Shallow draft                & $B+1$ (anchor skip)         & Frozen vocab token     & Bidirectional         & Frozen          \\
        \hybr\textbf{PTPv2 (Ours)} & \textbf{Shallow ($\sim$67M)} & \textbf{$B$ (all-position)} & \textbf{Learned mask}  & \textbf{Block-causal} & \textbf{Frozen} \\
        \bottomrule
    \end{tabular}
\end{table}

\paragraph{Decoupled Architecture and Accelerated Decoding.}
PTPv2 pairs the frozen 1B recognizer (the verifier) with a shallow, 4-layer decoder-only draft
module that shares the verifier's hidden width and output vocabulary head. Rather than predicting
from token embeddings alone, the draft fuses intermediate hidden states sampled across
shallow-to-deep verifier layers via a lightweight linear projection and normalization into
compact context features, which are injected directly into the key and value caches of every draft
layer.
Following the prefill stage, decoding proceeds in an iterative draft-and-verify loop: 
(1)~\textbf{Parallel Drafting:} Conditioned on the fused verifier context, the draft module takes the 
latest committed token embedding appended with a short block of learned mask queries, executing a single 
parallel forward pass to propose multiple future tokens simultaneously.
(2)~\textbf{Parallel Verification:} The frozen 1B verifier evaluates all drafted tokens in one forward pass. 
Because processing a short candidate block remains memory-bandwidth-bound, this parallel verification incurs 
nearly the same latency as a single autoregressive (AR) step. The verifier then accepts the longest matching 
prefix plus one correction token (or a bonus token if all proposals match). Finally, it extracts context 
features for the newly committed tokens and rolls back the rejected KV-cache entries. 
With the verifier frozen, greedy verification guarantees lossless, token-for-token equivalence with 
standard AR decoding while achieving speedups proportional to the accepted prefix length.

\paragraph{Design Advances over PTPv1 and DFlash.}
PTPv2 bridges feature-conditioned parallel drafting with causal next-token prediction (NTP), 
resolving key bottlenecks of coupled multi-token prediction (PTPv1) and block-diffusion drafting (DFlash) 
in structured document parsing:
\begin{itemize}
    \item \textbf{Decoupled drafting with an intact verifier (vs.\ PTPv1):} Rather than inserting register tokens into a fine-tuned target backbone (PTPv1), PTPv2 offloads drafting to
     a 4-layer side network. This reduces drafting compute by an order of magnitude while keeping the verifier 
     strictly frozen and lossless.
     \item \textbf{Zero-waste NTP alignment (vs.\ DFlash):} 
    DFlash leaves the anchor slot (slot 0) unsupervised, requiring $B+1$ query slots to emit $B$ proposals (yielding only $B-1$ proposals from $B$ queries).
     PTPv2 enforces standard NTP alignment across all positions---the anchor predicts the first future token and 
     mask slots predict subsequent ones---eliminating wasted query capacity.
    \item \textbf{Dedicated mask embedding (vs.\ DFlash):} Instead of repurposing a frozen vocabulary 
    token as the mask placeholder, PTPv2 jointly trains a standalone mask embedding, 
    preventing semantic interference with specialized document syntax (e.g., LaTeX commands and OTSL tags).
    \item \textbf{Causal intra-block attention (vs.\ DFlash):} Replacing DFlash's bidirectional diffusion 
    attention with strictly causal masking preserves left-to-right syntactic dependencies 
    (e.g., matching braces and table tags) and integrates seamlessly with standard causal KV-cache 
    kernels in vLLM.
\end{itemize}

\paragraph{Draft Training.}
We train the draft module from scratch on the Stage~3 dataset. During training, 
the 1B verifier, token embeddings, and LM head remain frozen; we only update the 4-layer draft decoder, 
context fusion projection, and learned mask embedding. All query slots are supervised using the frozen 
verifier's argmax predictions via D-PACE prefix-weighted cross-entropy loss~\citep{D_PACE_2026}.

\section{Experiments}
\label{sec:experiments}

\subsection{Evaluation Setup}
\label{sec:integrity}

Full-page parsing quality is evaluated on
OmniDocBench~v1.6~\citep{OmniDocBench_2025,MinerU2_5_Pro_2026} across text, isolated and display
formulas, tables, and reading order. Text and reading-order edit distances~\citep{Levenshtein_1966}
range from $0$ to $1$ (lower is better), while formula CDM~\citep{CDM_2024}, table
TEDS~\citep{PubTabNet_TEDS_2020}, and structure-only TEDS ($\mathrm{TEDS}_S$) are reported as
percentages (higher is better). Overall accuracy follows the benchmark definition:
\begin{equation}
    S_{\mathrm{overall}}=\frac{100\times(1-E_{\mathrm{text}})+S_{\mathrm{formula}}+S_{\mathrm{table}}}{3}.
    \label{eq:overall}
\end{equation}

\begin{table}[t]
    \centering
    \caption{
        \textbf{Comparison of document parsing results on OmniDocBench v1.6.} Nominal model sizes exclude
        external layout detectors and optional draft modules. By default, our pipeline (Stages~2--4) pairs the
        1B recognizer with PP-DocLayoutV3~\cite{PP_DocLayoutV3_2026}; we also report Stage~4 paired with
        Tele-layout~\cite{TeleOCR_2026} to examine sensitivity to upstream layout segmentation. } \label{tab:OmniDocBench_results}
    \fontsize{8.5}{10.5}\selectfont \setlength{\tabcolsep}{2.5pt}
    \begin{adjustbox}{max width=\linewidth}
        \begin{tabular}{@{} >{\centering\arraybackslash}p{0.080\linewidth} p{0.378\linewidth} >{\centering\arraybackslash}p{0.090\linewidth} *{6}{>{\centering\arraybackslash}p{0.064\linewidth}} @{}}
            \toprule
            \scriptsize\textbf{\makecell{Model \nextline Type}}            &
            \scriptsize\textbf{Model}                                      &
            \scriptsize\textbf{Size}                                       &
            \scriptsize\textbf{\makecell{Overall\nextline$\uparrow$}}      &
            \scriptsize\textbf{\makecell{Text\nextline Edit$\downarrow$}}  &
            \scriptsize\textbf{\makecell{Formula\nextline CDM$\uparrow$}}  &
            \scriptsize\textbf{\makecell{Table\nextline TEDS$\uparrow$}}   &
            \scriptsize\textbf{\makecell{Table\nextline TEDS$_S\uparrow$}} &
            \scriptsize\textbf{\makecell{Order\nextline Edit$\downarrow$}}
            \\

            \midrule
            \multirow{5}{*}[-1.8mm]
            {\scriptsize\makecell{General \nextline VLMs}}
                                                                           & \orglogo{\includegraphics[height=1.2em]{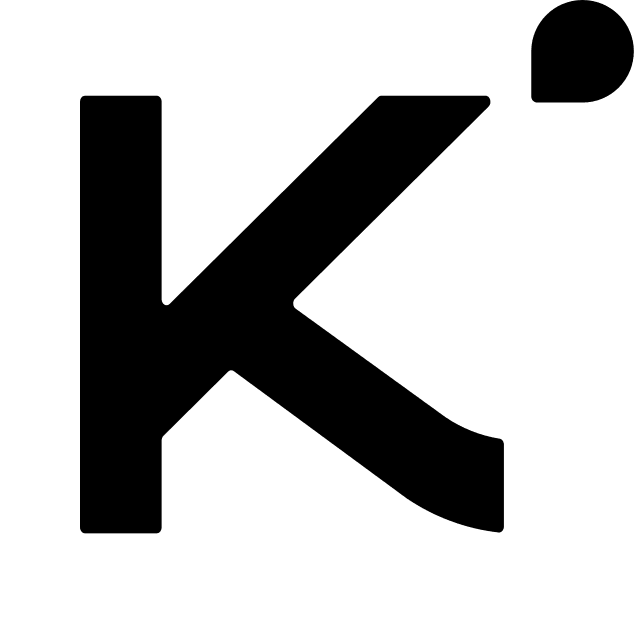}}Kimi K2.5~\cite{Kimi_K2_5_2026}                    & 1T        & 84.53          & 0.107          & 83.50          & 80.76          & 84.00          & 0.211          \\
                                                                           & \orglogo{\includegraphics[height=1.2em]{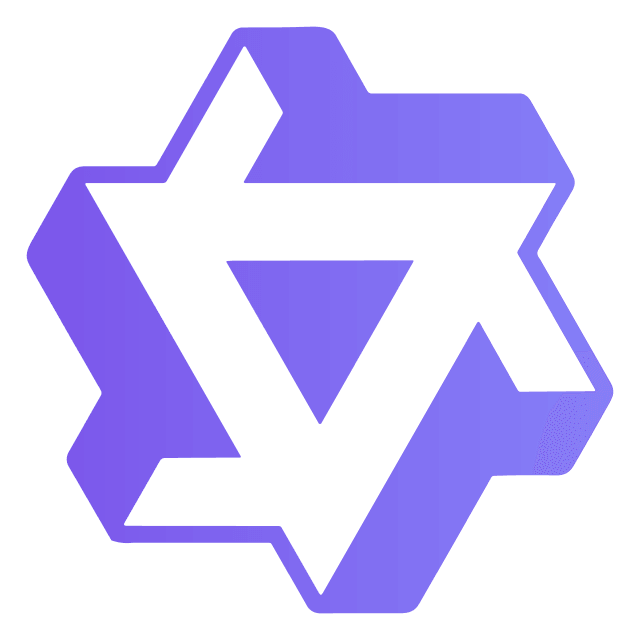}}Qwen3-VL-235B~\cite{Qwen3_VL_2025}                 & 235B-A22B & 89.78          & 0.063          & 92.55          & 83.07          & 86.75          & 0.166          \\
                                                                           & \orglogo{\includegraphics[height=1.2em]{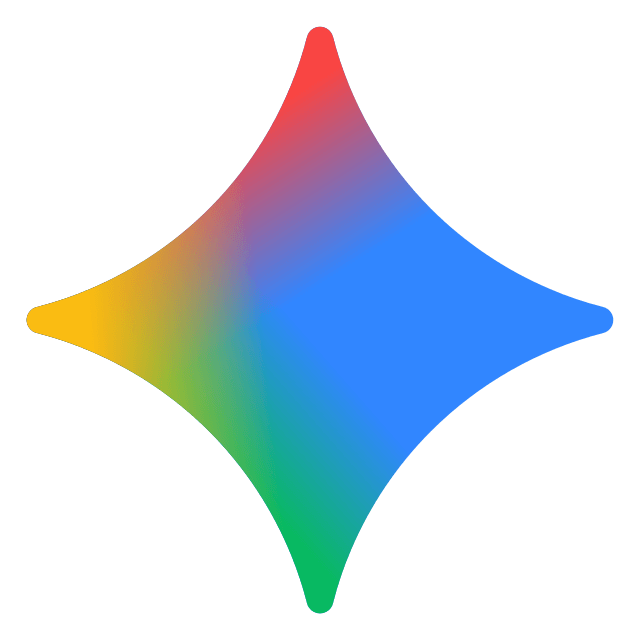}}Gemini 3 Flash~\cite{Gemini_3_Flash_2026}        & -         & 92.62          & 0.066          & 95.16          & 89.29          & 93.51          & 0.172          \\
                                                                           & \orglogo{\includegraphics[height=1.2em]{logo/org/gemini.png}}Gemini 3 Pro~\cite{Gemini_3_2026}                & -         & 92.91          & 0.064          & 95.99          & 89.15          & 92.96          & 0.165          \\
                                                                           & \orglogo{\includegraphics[height=1.2em]{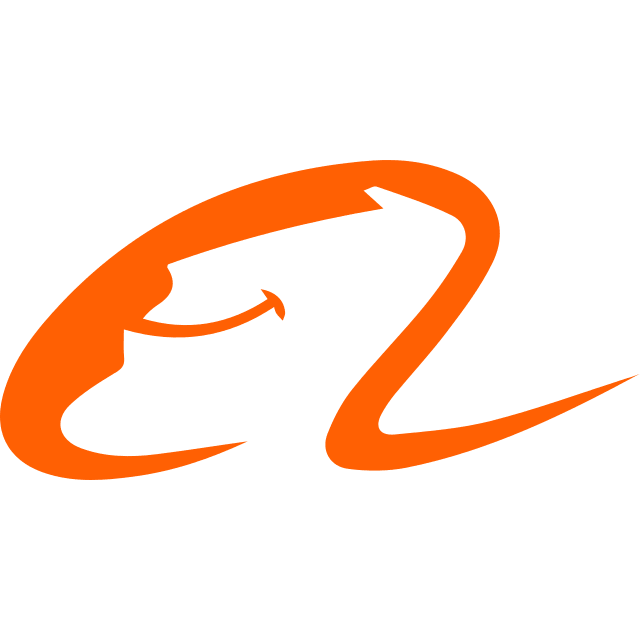}}Ovis2.6-30B-A3B~\cite{Ovis2_6_2026}             & 30B-A3B   & 93.70          & 0.035          & 95.17          & 89.44          & 92.40          & 0.135          \\
            \cmidrule(lr){1-9}
            \multirow{13}{*}[-2.8mm]
            {\scriptsize\makecell{Specialized \nextline VLMs}}
                                                                           & \orglogo{\includegraphics[height=1.2em]{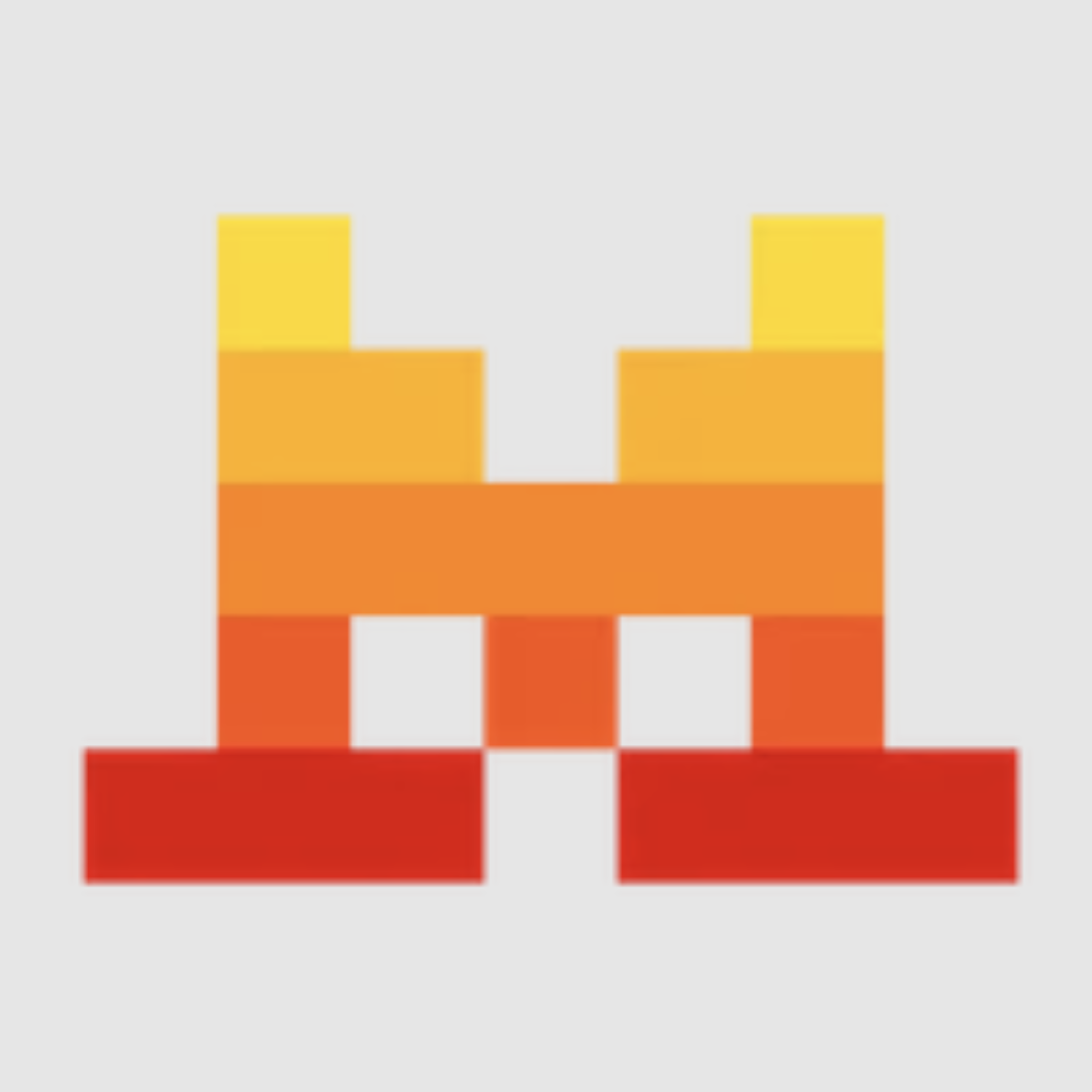}}Mistral OCR~\cite{Mistral_OCR_2025}           & -         & 85.66          & 0.097          & 89.91          & 76.78          & 80.93          & 0.171          \\
                                                                           & \orglogo{\includegraphics[height=1.2em]{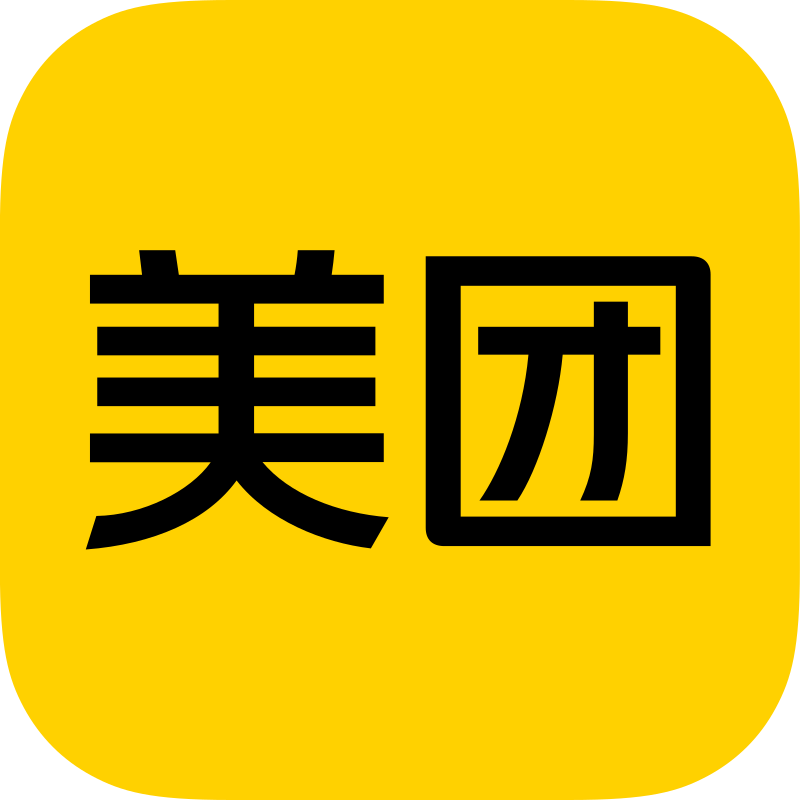}}OCRVerse~\cite{OCRVerse_2026}                   & 4B        & 88.60          & 0.063          & 89.61          & 82.44          & 86.27          & 0.163          \\
                                                                           & \orglogo{\includegraphics[height=1.2em]{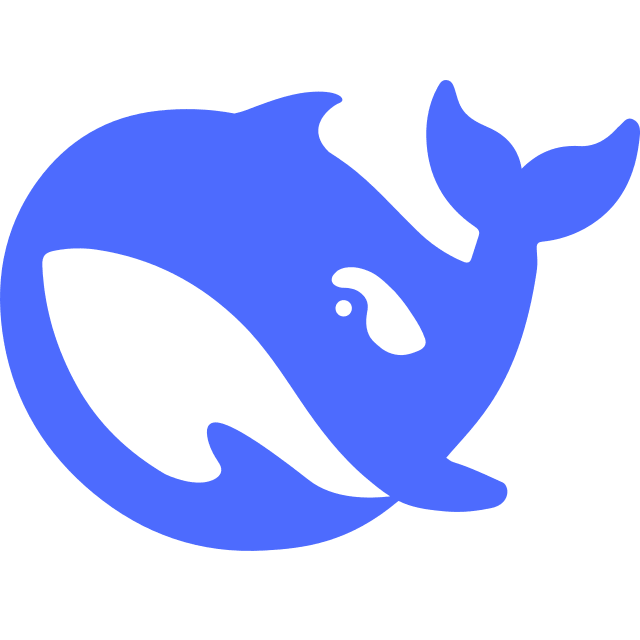}}DeepSeek-OCR 2~\cite{DeepSeek_OCR_2_2026}      & 3B-A0.5B  & 90.25          & 0.050          & 91.84          & 83.89          & 87.75          & 0.144          \\
                                                                           & \orglogo{\includegraphics[height=1.2em]{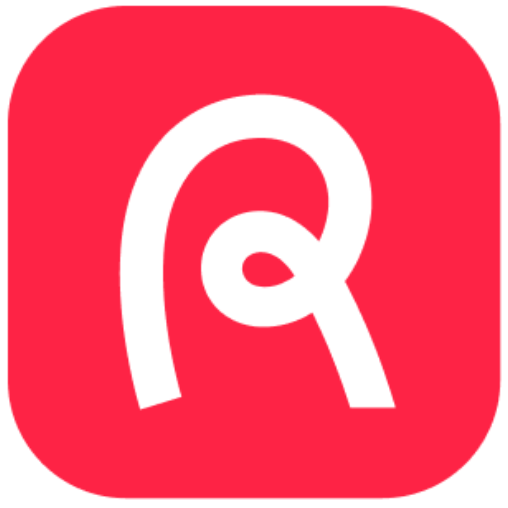}}dots.ocr~\cite{dots_ocr_2025}               & 3B        & 90.77          & 0.048          & 89.95          & 87.18          & 90.58          & 0.138          \\
                                                                           & \orglogo{\includegraphics[height=1.2em]{logo/org/xiaohongshu.png}}FireRed-OCR~\cite{FireRed_OCR_2026}         & 2B        & 93.26          & 0.037          & 95.44          & 88.04          & 91.06          & 0.131          \\
                                                                           & \orglogo{\includegraphics[height=1.2em]{logo/org/alibaba.png}}Logics-Parsing-v2~\cite{Logics_Parsing_v2_2026} & 4B        & 93.33          & 0.041          & 95.65          & 88.42          & 91.98          & 0.137          \\
                                                                           & \orglogo{\includegraphics[height=1.2em]{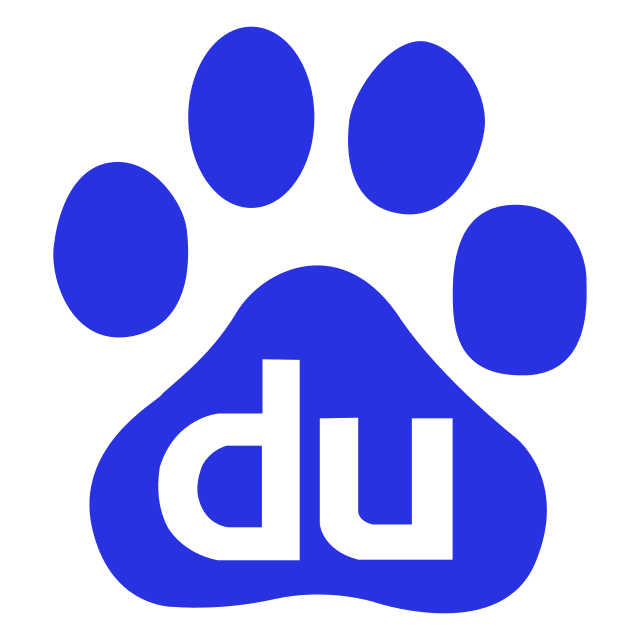}}Unlimited-OCR~\cite{Unlimited_OCR_2026}           & 3B-A0.5B  & 93.92          & 0.042          & 95.79          & 90.16          & 93.32          & 0.129          \\
                                                                           & \orglogo{\includegraphics[height=1.2em]{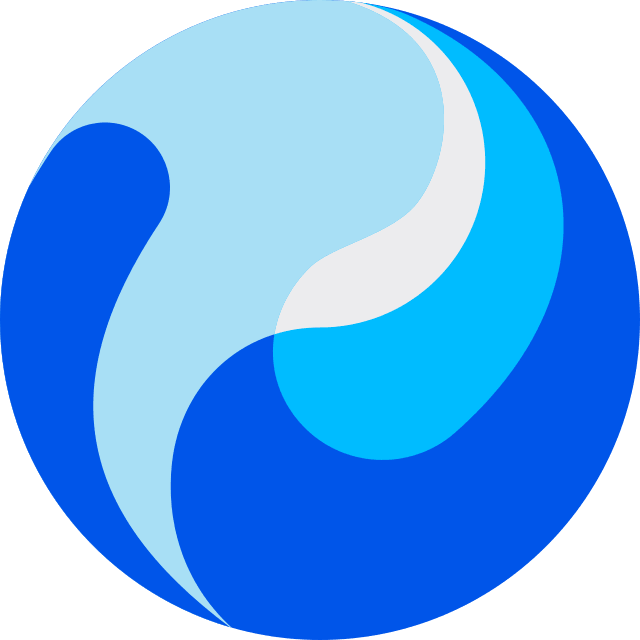}}HunyuanOCR-1.5~\cite{HunyuanOCR_1_5_2026}       & 1B        & 94.74          & 0.039          & 94.50          & 93.67          & 94.71          & 0.129          \\
                                                                           & \orglogo{\includegraphics[height=1.2em]{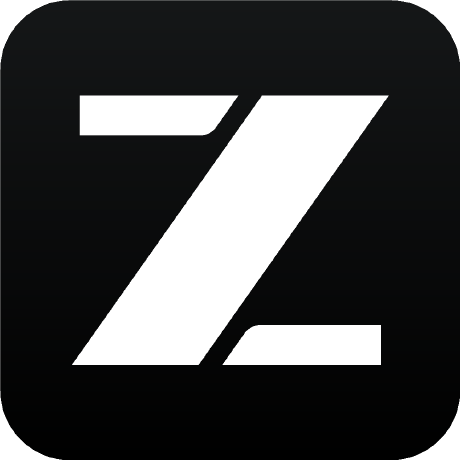}}GLM-OCR~\cite{GLM_OCR_2026}                       & 0.9B      & 95.22          & 0.044          & 97.18          & 92.83          & 95.39          & 0.133          \\
                                                                           & \orglogo{\includegraphics[height=1.2em]{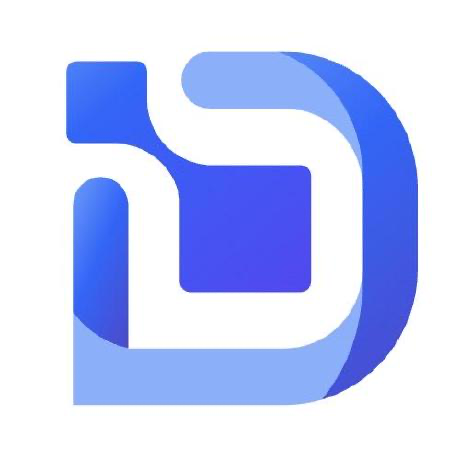}}MinerU2.5-Pro~\cite{MinerU2_5_Pro_2026}     & 1.2B      & 95.75          & 0.036          & 97.45          & 93.42          & 95.92          & 0.120          \\
                                                                           & \orglogo{\includegraphics[height=1.2em]{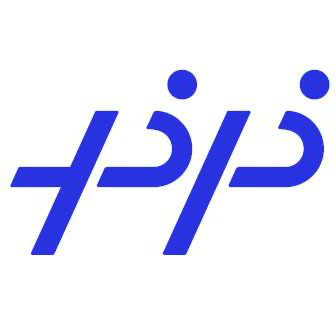}}PaddleOCR-VL-1.6~\cite{PaddleOCR_VL_1_6_2026}    & 0.9B      & 96.33          & 0.033          & 97.49          & 94.76          & 97.11          & 0.127          \\
                                                                           & \orglogo{\includegraphics[height=1.2em]{logo/org/alibaba.png}}OvisOCR2~\cite{OvisOCR2_2026}                   & 0.8B      & 96.58          & \textbf{0.025} & 97.53          & 94.76          & 97.16          & \textbf{0.111} \\
                                                                           & \orglogo{\includegraphics[height=1.2em]{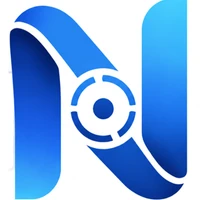}}TeleOCR~\cite{TeleOCR_2026}                      & 1.2B      & \textbf{96.91} & 0.027          & 96.59          & \textbf{96.82} & \textbf{98.18} & 0.118          \\
            \cmidrule(lr){1-9}
            \hybr\cellcolor{white}                                         & \orglogo{\includegraphics[height=1.2em]{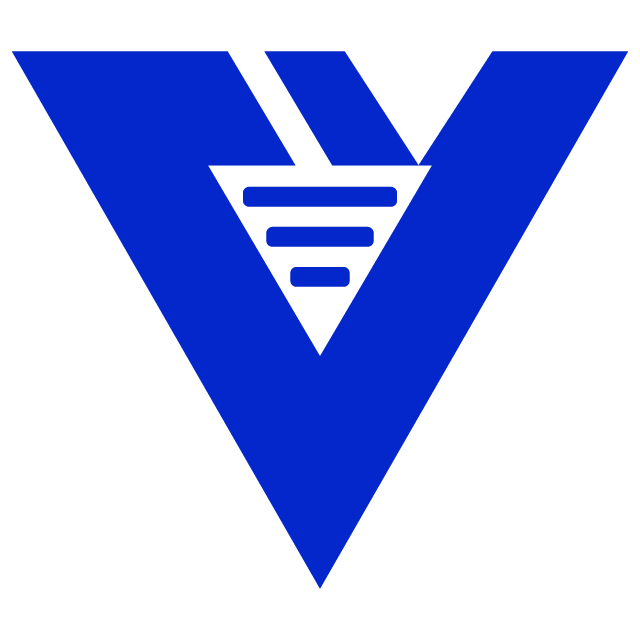}}VenusOCR 2.0 (Stage2)          & 1B        & 94.96          & 0.035          & 95.52          & 92.83          & 95.31          & 0.131          \\
            \hybr\cellcolor{white}                                         & \orglogo{\includegraphics[height=1.2em]{logo/org/venusocr.png}}VenusOCR 2.0 (Stage3)          & 1B        & 95.69          & 0.034          & 97.76          & 92.73          & 95.20          & 0.128          \\
            \hybr\cellcolor{white}                                         & \orglogo{\includegraphics[height=1.2em]{logo/org/venusocr.png}}VenusOCR 2.0 (Stage4 + PP-DocLayoutV3)          & 1B        & 96.24          & 0.033          & 97.86          & 94.16          & 96.32          & 0.128          \\
            \hybr\cellcolor{white}\multirow{-4}{*}{\scriptsize\makecell{Ours}}
                                                                           & \orglogo{\includegraphics[height=1.2em]{logo/org/venusocr.png}}VenusOCR 2.0 (Stage4 + Tele-layout)             & 1B        & 96.45          & 0.028          & \textbf{98.57} & 93.58          & 95.60          & 0.116          \\

            \bottomrule
        \end{tabular}
    \end{adjustbox}
\end{table}

Because VenusOCR~2.0 operates as a region-level recognizer within a layout-decoupled pipeline
(Section~\ref{sec:backbone}), full-page evaluation proceeds in two stages: an external layout detector
first segments each document page into categorized regions (text, formulas, and tables) and predicts
reading order, after which VenusOCR~2.0 transcribes each cropped region independently before assembling
the page-level output. By default, all stage checkpoints (Stages~2--4) use
PP-DocLayoutV3~\citep{PP_DocLayoutV3_2026}, a real-time 33M-parameter detector that jointly predicts
region boundaries, element categories, and reading order in a single forward pass. To isolate the
sensitivity of end-to-end parsing to upstream layout segmentation and reading-order quality, we also
evaluate the Stage~4 recognizer paired with Tele-layout~\citep{TeleOCR_2026}, the deformation-aware
layout detector from TeleOCR. Throughout Table~\ref{tab:OmniDocBench_results}, nominal model sizes
reflect the 1B recognition backbone only and exclude external layout detectors as well as optional
PTPv2 draft modules.

\subsection{Main Results}
\label{sec:progression}

As shown in Table~\ref{tab:OmniDocBench_results}, our VenusOCR~2.0 (Stage~4 + PP-DocLayoutV3) achieves
an overall score of 96.24 on OmniDocBench~v1.6, which further rises to 96.45 when paired with
Tele-layout---notably setting a new state of the art in formula recognition (98.57 CDM). Despite its
compact 1B footprint, VenusOCR~2.0 substantially outperforms massive general-purpose VLMs (e.g.,
Gemini~3~Pro~\citep{Gemini_3_2026} and Qwen3-VL-235B~\citep{Qwen3_VL_2025}) and remains highly
competitive with leading specialized parsers~\citep{PaddleOCR_VL_1_6_2026, OvisOCR2_2026, TeleOCR_2026}.
\begin{table}[t]
    \centering
    \footnotesize
    \setlength{\tabcolsep}{4pt}
    \caption{Stage-by-stage progression on OmniDocBench~v1.6. Stages~2--4 use PP-DocLayoutV3 by default;
        the final row isolates switching the external layout detector at Stage~4. Positive similarity deltas
        and negative edit-distance deltas indicate gains.} \label{tab:progression}
    \begin{tabular}{@{}lrrrrr@{}}
        \toprule
        Transition                                       & $\Delta$Overall & $\Delta$Text edit & $\Delta$CDM & $\Delta$TEDS & $\Delta$Order edit \\
        \midrule
        Stage 2 $\rightarrow$ Stage 3 (Curated SFT)      & $+0.73$         & $-0.001$          & $+2.24$     & $-0.10$      & $-0.003$           \\
        Stage 3 $\rightarrow$ Stage 4 (Metric RL)        & $+0.55$         & $-0.001$          & $+0.10$     & $+1.43$      & $0.000$            \\
        Stage 4 (PP-DocLayoutV3 $\rightarrow$ Tele-layout) & $+0.21$       & $-0.005$          & $+0.71$     & $-0.58$      & $-0.012$           \\
        \bottomrule
    \end{tabular}
\end{table}
Table~\ref{tab:progression} highlights the complementary roles of our multi-stage training pipeline.
Starting from the Stage~2 pretrained baseline (94.96 with PP-DocLayoutV3), Stage~3 (Curated SFT) drives
a major leap in formula accuracy (+2.24 CDM) via consensus-filtered and synthetic data. Subsequently,
Stage~4 (Metric RL) specifically boosts table parsing (+1.43 TEDS) by directly optimizing structural
TEDS rewards, bringing the overall score to 96.24 under the default PP-DocLayoutV3 detector. Finally,
replacing PP-DocLayoutV3 with Tele-layout yields sharper region boundaries and reading order, further
unlocking the recognition capacity of VenusOCR~2.0 and lifting the overall score to 96.45 (with a slight
$-0.58$ TEDS trade-off due to table crop-boundary differences).

\subsection{Data Iteration and Representation Studies}
\label{sec:ablation}

\begin{table}[t]
    \centering
    \footnotesize
    \setlength{\tabcolsep}{4.5pt}
    \caption{Cumulative progression across RSI data and normalization cycles on OmniDocBench~v1.6.}
    \label{tab:rsi_cycles}
    \begin{tabular}{@{}lrrrrrr@{}}
        \toprule
        Iteration                                   & Overall$\uparrow$ & Text$\downarrow$ & Formula$\uparrow$ & Table$\uparrow$ & Table$_S$$\uparrow$ & Order$\downarrow$ \\
        \midrule
        Cycle 0 (Initial Pretrain Baseline)         & 92.92             & 0.038            & 93.85             & 88.71           & 92.01               & 0.132             \\
        Cycle 1 (Initial SFT + Bad-Case Synthesis)  & 94.18             & 0.036            & 95.87             & 90.27           & 93.51               & 0.129             \\
        Cycle 2 (Corpus Rebalance + Format Cleanup) & 94.77             & 0.034            & 96.55             & 91.16           & 93.61               & 0.129             \\
        Cycle 3 (OTSL + Format Refinement)          & 95.48             & 0.035            & 97.63             & 92.33           & 94.96               & 0.128             \\
        Cycle 4 (Final Stage~3 Curated SFT)         & 95.69             & 0.034            & 97.76             & 92.73           & 95.20               & 0.128             \\
        \bottomrule
    \end{tabular}
\end{table}

\paragraph{Iterative RSI Progression.}
Table~\ref{tab:rsi_cycles} summarizes how recognition quality evolves across successive RSI data
iterations on OmniDocBench~v1.6. Progressing from the initial pretraining baseline
through targeted synthesis, data rebalancing, and format normalization steadily boosts overall
parsing accuracy, particularly on formulas and tables. 
Notably, strict quality control proved essential during this process: directly mixing in unfiltered synthetic tables 
and formulas in an intermediate cycle degraded performance, and accuracy only recovered after we filtered 
out noisy markup and rebalanced the data mixture. Note that because later cycles update both data composition 
and formatting rules simultaneously, Table~\ref{tab:rsi_cycles} reflects cumulative system gains rather than 
isolated single-factor ablations.

\begin{table}[t]
    \centering
    \footnotesize
    \setlength{\tabcolsep}{4.5pt}
    \caption{Development comparisons on table serialization and vocabulary extension (OmniDocBench~v1.6).
        The upper block compares HTML and OTSL table formats under updated mixture weighting; the lower
        block compares extending the vocabulary with OTSL tokens alone versus adding LaTeX keywords.}
    \label{tab:repr_ablation}
    \begin{tabular}{@{}lrrrrrr@{}}
        \toprule
        Configuration                              & Overall$\uparrow$ & Text$\downarrow$ & Formula$\uparrow$ & Table$\uparrow$ & Table$_S$$\uparrow$ & Order$\downarrow$ \\
        \midrule
        HTML Table Format (Rebalanced)             & 94.77             & 0.034            & 96.55             & 91.16           & 93.61               & 0.129             \\
        OTSL Table Format (Rebalanced)             & 95.27             & 0.035            & 96.51             & 92.79           & 95.17               & 0.129             \\
        \midrule
        OTSL + LaTeX Keyword Tokens                & 95.16             & 0.036            & 97.23             & 91.85           & 94.66               & 0.128             \\
        OTSL Tokens Only                           & 95.39             & 0.036            & 97.45             & 92.33           & 94.96               & 0.128             \\
        \bottomrule
    \end{tabular}
\end{table}

\paragraph{Table Serialization and Vocabulary Extension.}
Table~\ref{tab:repr_ablation} evaluates two structural representation choices: using atomic OTSL tokens for 
tables and expanding the vocabulary for LaTeX. First, replacing verbose HTML table markup with compact OTSL tokens (alongside updated data weights) 
substantially improves table recognition and topology without affecting text or formula accuracy. 
In contrast, further expanding the decoder vocabulary with common LaTeX keywords slightly 
degrades both formula and table scores. This suggests that while OTSL introduces genuinely compact primitives 
that shorten table sequences and reduce syntax errors, common LaTeX commands are already well tokenized by the
pretrained subword vocabulary.

\subsection{Inference Acceleration with PTPv2}
\label{sec:ptpv2_results}
\label{sec:ptpv2_runtime}

We evaluate PTPv2 from offline draft training through live vLLM serving. The 4-layer draft
($\sim$67M trainable parameters, $B=5$ proposals, random initialization) is trained for five epochs
on a $95/5$ split with the verifier frozen.

\begin{figure}[t]
    \centering
    \includegraphics[width=0.86\linewidth]{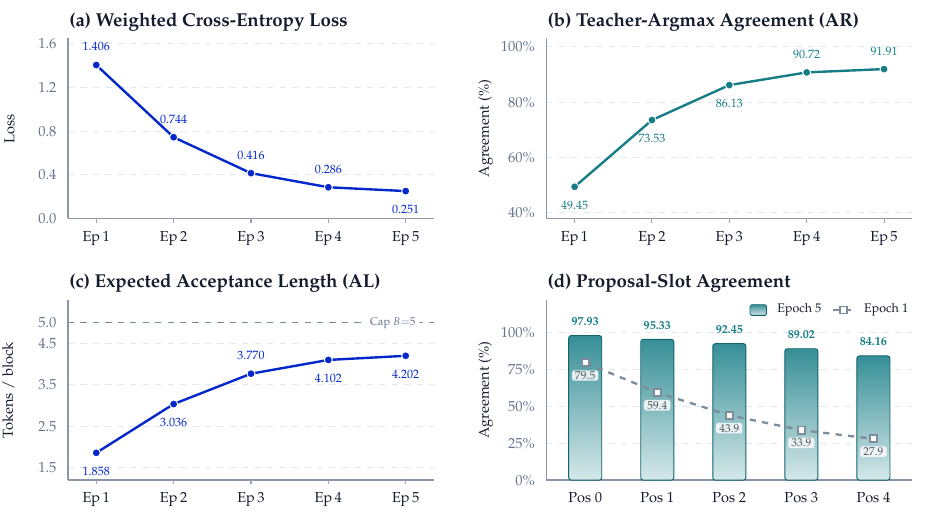}
    \caption{PTPv2 offline validation convergence across training epochs: (a)~weighted cross-entropy loss,
        (b)~mean teacher-argmax agreement, (c)~expected acceptance length proxy relative to the proposal cap,
        and (d)~per-position agreement from Epoch~1 to Epoch~5.}
    \label{fig:ptpv2_training}
\end{figure}

\paragraph{Offline proposal learning.}
Figure~\ref{fig:ptpv2_training} illustrates the validation convergence of the draft module over five 
training epochs. As the D-PACE loss steadily decreases, overall top-1 agreement with the frozen verifier
climbs from $49.45\%$ to $91.91\%$, driving the offline expected acceptance length (EAL) proxy---distinct from online engine acceptance---from $1.86$ 
to $4.20$ (out of a block size of $B=5$). Notably, accuracy remains high across all proposal positions: 
by Epoch~5, position-wise agreement ranges from $97.93\%$ at the initial slot to $84.16\%$ at the final 
slot (position~4), with later positions improving by over $50$ percentage points during training.

\paragraph{Serving benchmark.}
We benchmark standard autoregressive (AR) and PTPv2 decoding in vLLM~\citep{vLLM_2023} 
(BF16 with FlashAttention~\citep{FlashAttention2_2023} and CUDA graphs) on a single NVIDIA H20 (96GB) 
GPU across concurrency levels $c \in \{1, 4, 8, 16, 32\}$. The evaluation workload comprises all $29{,}849$ region crops 
(text, formulas, and tables) from the $1{,}651$ pages of OmniDocBench~v1.6. 
We measure end-to-end recognition latency and throughput 
from client request to stream completion, excluding layout detection and post-processing.

\begin{figure}[t]
    \centering
    \includegraphics[width=0.86\linewidth]{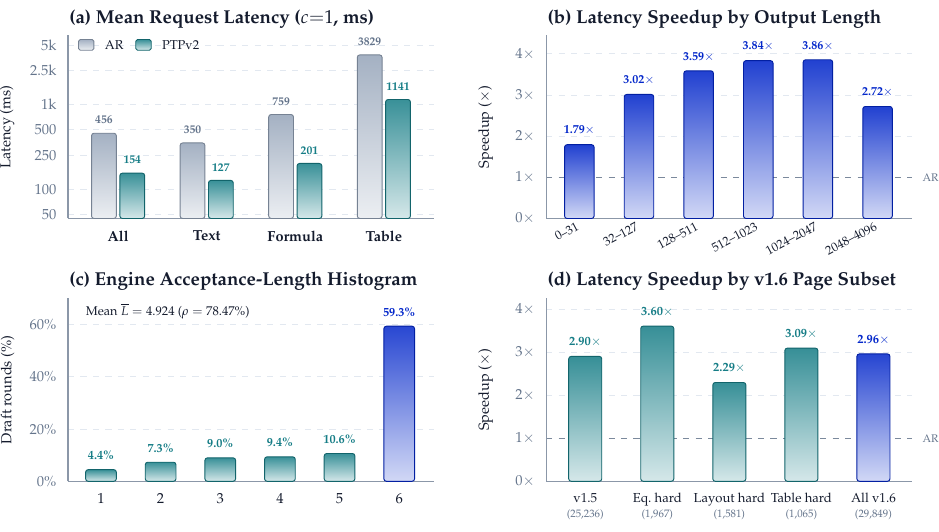}
    \caption{Single-request ($c=1$) latency and acceptance breakdown on OmniDocBench~v1.6: (a)~mean latency
        by task, (b)~speedup by output length, (c)~acceptance-length distribution per draft round, and
        (d)~speedup across page subsets.}
    \label{fig:ptpv2_single_request_breakdown}
\end{figure}

\paragraph{Single-Request Latency and Task Breakdown.}
Figure~\ref{fig:ptpv2_single_request_breakdown} summarizes single-request ($c=1$) performance. 
Overall, PTPv2 reduces mean latency by $2.96\times$ (from $456.1$ to $154.3\text{ ms}$) and boosts 
wall-clock throughput from $128.5$ to $379.1\text{ tokens/s}$. Speedups scale with sequence length and structural 
regularity: while standard text improves by $2.77\times$, longer and more structured regions achieve even 
larger gains—reaching $3.36\times$ on tables ($392.4\text{ tokens/s}$ per-request rate) and $3.78\times$ on isolated formulas ($554.0\text{ tokens/s}$ per-request rate). 
This trend holds across challenging subsets, with \texttt{equation\_hard} and \texttt{table\_hard} accelerating 
by $3.60\times$ and $3.09\times$, respectively. At the system level, while median time-to-first-token (TTFT) remains 
comparable ($40.3$ vs.\ $39.1\text{ ms}$), PTPv2 cuts median time-per-output-token (TPOT) by 
$3.65\times$ ($6.40\to 1.75\text{ ms}$), reducing P95 latency by over $3.4\times$.

\begin{table}[t]
    \centering
    \footnotesize
    \setlength{\tabcolsep}{3pt}
    \caption{Single-GPU H20 concurrency scaling on OmniDocBench~v1.6. $\rho$ and $\tau$ denote online
        draft-token acceptance rate and mean tokens per step; all non-truncated requests match AR
        token-for-token.}
    \label{tab:ptpv2_runtime}
    \begin{tabular}{@{}rrrrrrrrrr@{}}
        \toprule
            & \multicolumn{2}{c}{Latency (ms)} & \multicolumn{2}{c}{Throughput (tok/s)} & \multicolumn{2}{c}{Speedup} &              &               &               \\
        $c$ & AR                               & PTPv2                                  & AR                          & PTPv2   & $S_L$ & $S_T$ & $\rho$ (\%) & $\tau$ \\
        \midrule
        1   & 456.14                           & 154.27                                 & 128.45                      & 379.10  & 2.957 & 2.951 & 78.47       & 4.924     \\
        4   & 594.84                           & 251.87                                 & 400.29                      & 944.68  & 2.362 & 2.360 & 78.21       & 4.910          \\
        8   & 739.14                           & 370.42                                 & 645.76                      & 1287.65 & 1.995 & 1.994 & 78.17       & 4.909       \\
        16  & 985.24                           & 599.66                                 & 969.48                      & 1591.64 & 1.643 & 1.642 & 78.23       & 4.911         \\
        32  & 1454.11                          & 1055.65                                & 1296.11                     & 1802.78 & 1.377 & 1.391 & 78.16       & 4.908      \\
        \bottomrule
    \end{tabular}
\end{table}

\begin{figure}[t]
    \centering
    \includegraphics[width=0.85\linewidth]{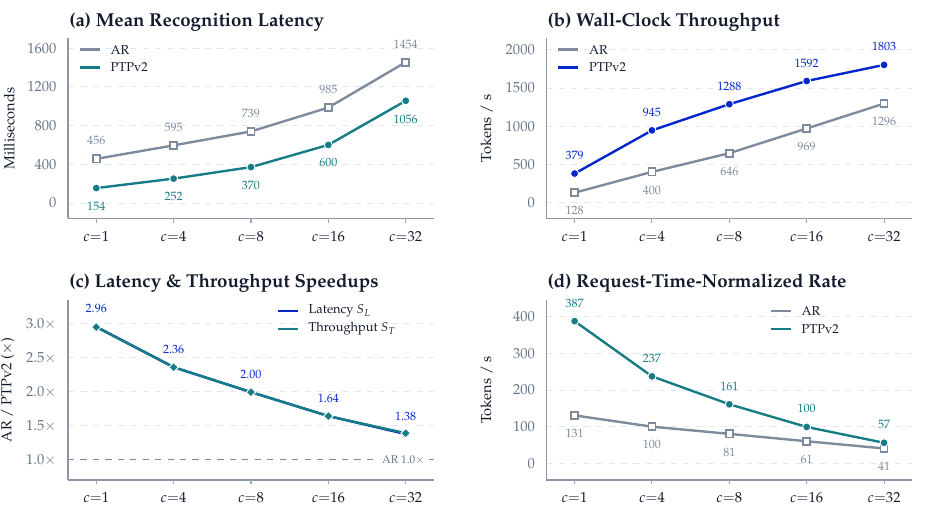}
    \caption{Concurrency scaling across $c=1$--$32$ comparing PTPv2 against autoregressive decoding on
        latency, throughput, speedup, and per-stream token rate.}
    \label{fig:ptpv2_concurrency_scaling}
\end{figure}

\paragraph{Concurrency Scaling and Lossless Equivalence.}
Table~\ref{tab:ptpv2_runtime} and Figure~\ref{fig:ptpv2_concurrency_scaling} show that PTPv2 consistently 
accelerates throughput across concurrency levels $c \in \{1, 4, 8, 16, 32\}$ by $2.95\times$ down to 
$1.39\times$, reaching $1{,}803\text{ tokens/s}$ at $c=32$ (vs.\ $1{,}296\text{ tokens/s}$ for AR). 
Notably, PTPv2 at $c=8$ matches the throughput of AR at $c=32$ ($\sim\!1{,}290\text{ tokens/s}$) while 
reducing mean latency by $3.9\times$ ($370$ vs.\ $1{,}454\text{ ms}$). Across all concurrencies, online 
draft acceptance remains stable at $\sim\!78.3\%$ ($\tau \approx 4.91\text{ tokens/step}$), 
confirming that the narrower relative speedup at high concurrency is due to increased compute saturation 
in batched AR rather than acceptance degradation. Finally, PTPv2 preserves exact token-for-token equivalence 
with AR on $100\%$ of non-truncated requests ($99.96\%$ overall, with minor divergence occurring only 
in $12$ degenerate repetitive samples that hit the $4{,}096$-token limit).
\section{Conclusion}

We presented VenusOCR~2.0, a compact 1B vision-language model for efficient and effective document parsing. 
By pairing a bounded RSI data engine with a progressive training pipeline, VenusOCR~2.0 achieves 
96.45 overall accuracy and a state-of-the-art 98.57 formula CDM on OmniDocBench~v1.6, outperforming massive 
general-purpose VLMs at a fraction of their size. To accelerate deployment, we introduced PTPv2, a lightweight 
($\sim$67M) causal parallel draft module that integrates seamlessly into vLLM. PTPv2 delivers a $2.96\times$ 
single-request speedup (reaching $3.78\times$ on formulas and $3.36\times$ on tables) and up to $2.36\times$ 
higher batched throughput while guaranteeing lossless, token-for-token equivalence with standard autoregressive
decoding.
\FloatBarrier

\bibliography{references}
\bibliographystyle{colm2026_conference}
\clearpage
\appendix

\section{Model Architecture and Training Configurations}
\label{app:configs}

Table~\ref{tab:architecture} summarizes the 1B recognition backbone. Input preprocessing aligns image
dimensions to multiples of 28 up to roughly 1.6M pixels, and the language decoder uses grouped-query
attention with 16 query heads and 8 key--value heads. Table~\ref{tab:referenceconfigs} lists the
learning rates and sequence budgets across stages. Packed pretraining preserves sample boundaries and
multimodal position indices within each container rather than treating the packed budget as a
single-document context length, while Stage~4 RL uses 16 rollouts per prompt with DAPO clipping and
overlong-completion masking.

\begin{table}[htbp]
    \centering
    \small
    \caption{Recognition backbone configuration. Shared input/output embeddings are counted once; external
        layout and draft modules are excluded.} \label{tab:architecture}
    \begin{tabular}{lll}
        \toprule
        Component        & Configuration                                          & Parameters \\
        \midrule
        Vision encoder   & 27 layers, width 1152, patch size 14                   & 413M       \\
        Projector        & $2\times2$ merge; $4608\rightarrow4608\rightarrow1024$ & 26M        \\
        Language decoder & 28 layers, width 1024, FFN width 3072                  & 596M       \\
        \midrule
        Total            & Shared input/output embeddings                         & $\sim$1B   \\
        \bottomrule
    \end{tabular}
\end{table}

\begin{table}[htbp]
    \centering
    \small
    \caption{Training hyperparameters across stages. Learning rates list vision / projector / decoder where
        distinct.} \label{tab:referenceconfigs}
    \begin{tabular}{L{0.23\linewidth}L{0.43\linewidth}L{0.24\linewidth}}
        \toprule
        Stage         & Learning rates                                          & Sequence setting \\
        \midrule
        1             & -- / $2\!\times\!10^{-4}$ / --                          & 4096             \\
        2             & $5\!\times\!10^{-6}$ / $5\!\times\!10^{-6}$ / $10^{-5}$ & 40960 packed     \\
        3             & $5\!\times\!10^{-6}$ for all components                 & 16384            \\
        4             & $5\!\times\!10^{-7}$                                    & 5120 completion  \\
        PTPv2         & $3\!\times\!10^{-4}$, draft only                        & 5 proposals      \\
        \bottomrule
    \end{tabular}
\end{table}

\section{PTPv2 Implementation Details}
\label{app:ptpv2}
\label{app:efficiency}

\paragraph{Context fusion and evaluation metrics.}
The draft module fuses selected intermediate hidden states from the frozen verifier through a linear
projection and RMS normalization, attending over a same-document prefix window of up to 2,048 tokens
prior to the current anchor. Offline validation tracks the marginal expected acceptance length (EAL)
proxy computed from per-position teacher agreement rates, which differs from true joint prefix
acceptance under draft-generated prefixes. Online vLLM serving instead logs exact engine counters:
draft acceptance rate $\rho$, accepted draft tokens per round, average tokens committed per step
$\tau$, latency speedup $S_L$, and wall-clock throughput speedup $S_T$.

\section{Detailed PTPv2 Runtime Breakdown}
\label{app:ptpv2_runtime_breakdown}

\subsection{Single-Request Breakdowns}

Tables~\ref{tab:ptpv2_output_length}--\ref{tab:ptpv2_language} break down the single-request H20
benchmark across output length and page language. Speedup grows steadily
with output length as parallel drafting amortizes prefill and per-step overhead, while even short
crops such as page numbers yield net gains. All non-truncated requests match autoregressive decoding
token-for-token.

\begin{table}[!htbp]
    \centering
    \small
    \setlength{\tabcolsep}{3pt}
    \caption{Single-request ($c=1$) results grouped by AR output length (latency in ms, request rate in
        tokens/s, speedup $S_L$).}
    \label{tab:ptpv2_output_length}
    \begin{tabular}{lrrrrrrr}
        \toprule
                      &         & \multicolumn{2}{c}{Latency (ms)} & \multicolumn{2}{c}{Request rate} &       &               \\
        Output tokens & Regions & AR                               & PTPv2                            & AR    & PTPv2 & $S_L$ & Exact match   \\
        \midrule
        0--31         & 15,962  & 108.40                           & 60.41                            & 106.45 & 191.00 & 1.794 & 15,962/15,962 \\
        32--127       & 10,659  & 486.45                           & 161.26                           & 138.90 & 419.01 & 3.017 & 10,659/10,659 \\
        128--511      & 2,996   & 1423.62                          & 396.46                           & 140.66 & 505.09 & 3.591 & 2,996/2,996   \\
        512--1023     & 152     & 5412.00                          & 1410.58                          & 128.15 & 491.67 & 3.837 & 152/152       \\
        1024--2047    & 56      & 11998.86                         & 3109.92                          & 115.38 & 445.15 & 3.858 & 56/56         \\
        2048--4096    & 24      & 39181.48                         & 14385.33                         & 102.76 & 279.88 & 2.724 & 12/24         \\
        \bottomrule
    \end{tabular}
\end{table}

\begin{table}[!htbp]
    \centering
    \small
    \setlength{\tabcolsep}{3pt}
    \caption{Single-request ($c=1$) results grouped by page language.}
    \label{tab:ptpv2_language}
    \begin{tabular}{lrrrrrrr}
        \toprule
                              &         & \multicolumn{2}{c}{Latency (ms)} & \multicolumn{2}{c}{Request rate} &       &               \\
        Page language         & Regions & AR                               & PTPv2                            & AR    & PTPv2 & $S_L$ & Exact match   \\
        \midrule
        English               & 13,275  & 527.20                           & 170.85                           & 131.52 & 405.84 & 3.086 & 13,266/13,275 \\
        Simplified Chinese    & 14,451  & 418.82                           & 145.53                           & 131.97 & 379.80 & 2.878 & 14,449/14,451 \\
        Mixed English/Chinese & 1,902   & 256.88                           & 108.30                           & 113.61 & 269.46 & 2.372 & 1,901/1,902   \\
        Traditional Chinese   & 188     & 380.39                           & 134.65                           & 122.03 & 344.75 & 2.825 & 188/188       \\
        Other                 & 33      & 131.61                           & 73.28                            & 114.89 & 206.34 & 1.796 & 33/33         \\
        \bottomrule
    \end{tabular}
\end{table}

\FloatBarrier
\clearpage
\subsection{Concurrency Latency and Acceptance}
\label{app:ptpv2_concurrency_plots}

\begin{figure}[t]
    \centering
    \includegraphics[width=0.86\linewidth]{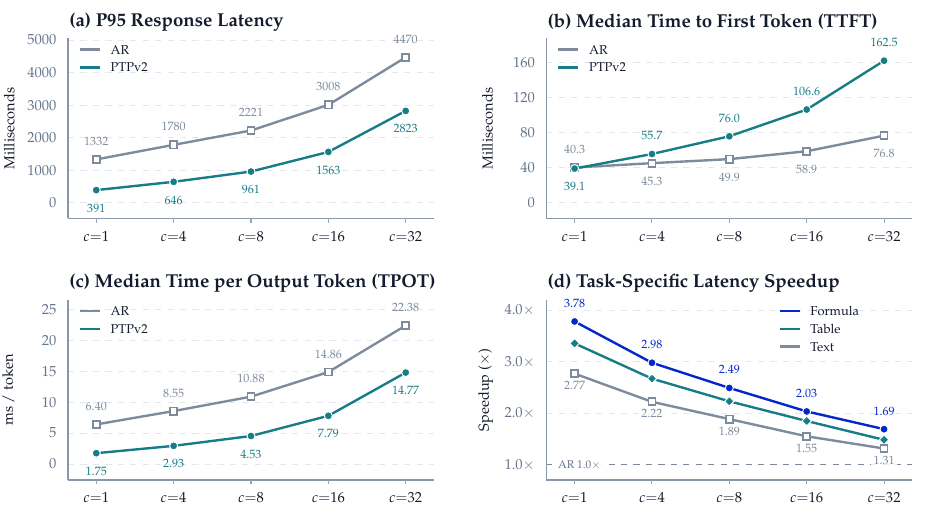}
    \caption{Detailed concurrency breakdown across $c=1$--$32$: P95 latency, median TTFT, median TPOT, and
        task-specific speedups.}
    \label{fig:ptpv2_concurrency_latency}
\end{figure}

\begin{figure}[t]
    \centering
    \includegraphics[width=0.86\linewidth]{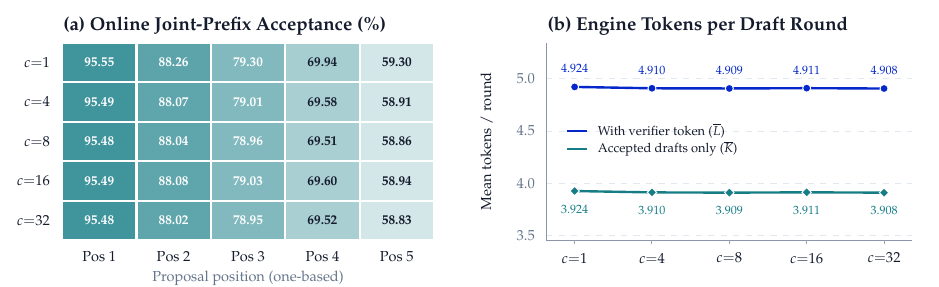}
    \caption{Online draft acceptance across concurrency levels ($c=1$--$32$): joint-prefix acceptance rates
        by position (left) and mean accepted tokens per round (right).}
    \label{fig:ptpv2_concurrency_acceptance}
\end{figure}

Across concurrency levels $c=1\text{--}32$, PTPv2 roughly halves P95 tail latency and sharply reduces
median time-per-output-token (TPOT) while keeping first-token latency comparable to AR
(Figure~\ref{fig:ptpv2_concurrency_latency}). Online joint-prefix acceptance remains invariant across all
batch sizes (Figure~\ref{fig:ptpv2_concurrency_acceptance}), confirming that narrower speedups at high
concurrency reflect compute saturation in batched AR rather than draft degradation.

\FloatBarrier
\clearpage
\section{Qualitative Case Studies}
\label{app:case_studies}

\begin{figure}[t]
    \centering
    \includegraphics[width=0.84\linewidth]{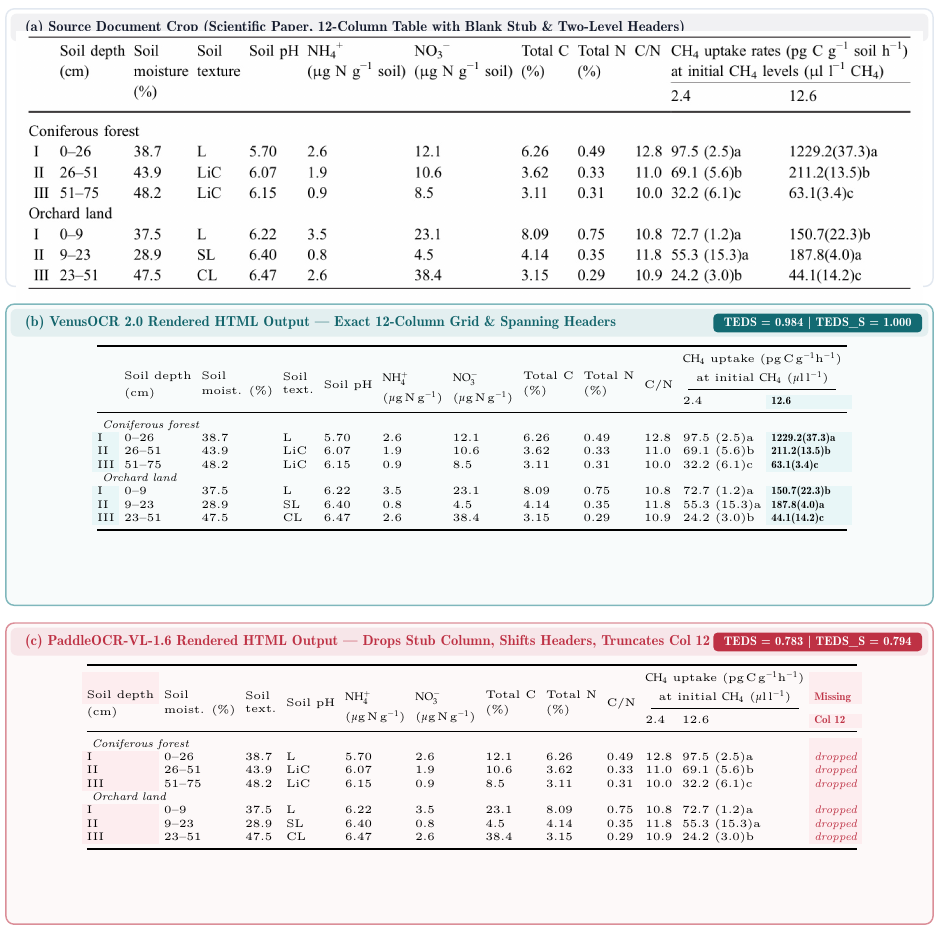}
    \caption{Qualitative comparison on a multi-level scientific table.}
    \label{fig:case_table_sci}
\end{figure}

\begin{figure}[t]
    \centering
    \includegraphics[width=0.84\linewidth]{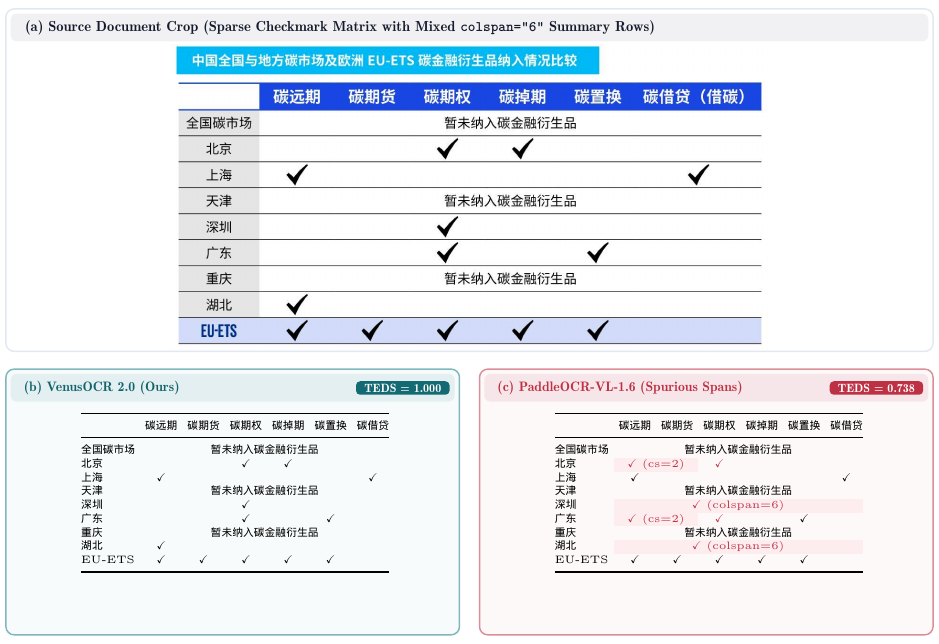}
    \caption{Qualitative comparison on a sparse checkmark matrix table.}
    \label{fig:case_table_matrix}
\end{figure}

\begin{figure}[t]
    \centering
    \includegraphics[width=0.76\linewidth]{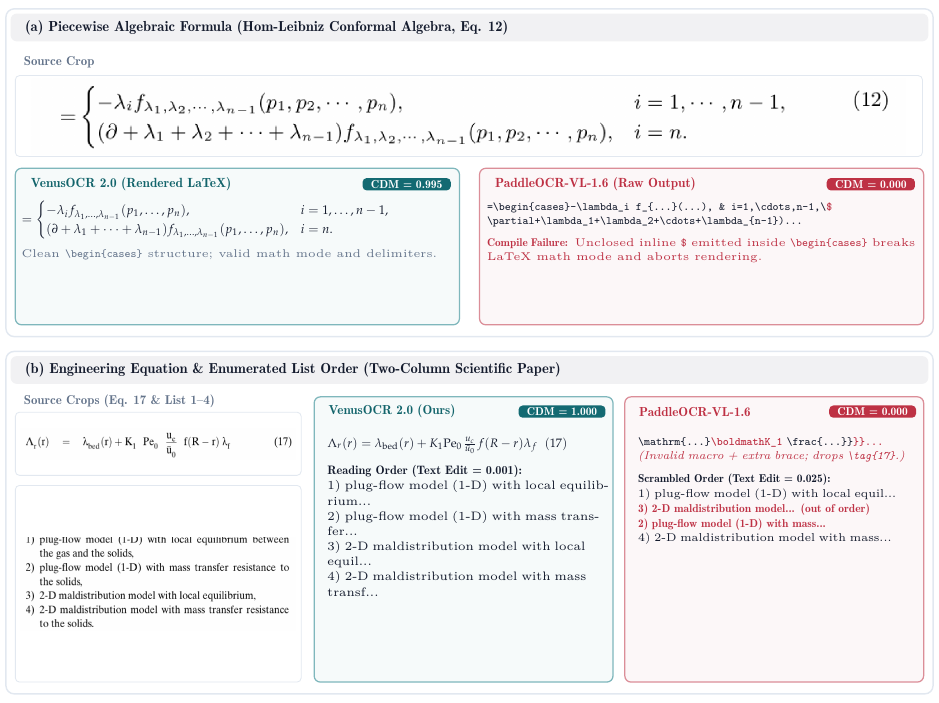}
    \caption{Qualitative comparison on mathematical formulas and reading order.}
    \label{fig:case_formula}
\end{figure}

\begin{figure}[t]
    \centering
    \includegraphics[width=0.76\linewidth]{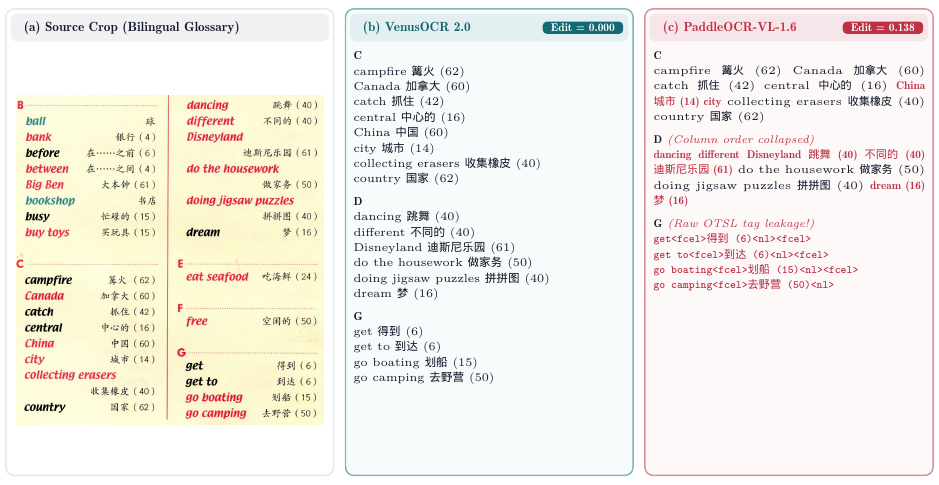}
    \caption{Qualitative comparison on a two-column bilingual layout.}
    \label{fig:case_text_bilingual}
\end{figure}

Figures~\ref{fig:case_table_sci}--\ref{fig:case_text_bilingual} illustrate representative parsing comparisons
against PaddleOCR-VL-1.6 across complex tables, mathematical formulas, and multi-column layouts.

\clearpage

On the sparse checkmark matrix table with mixed \texttt{colspan="6"} summary rows
(Figure~\ref{fig:case_table_matrix}), atomic OTSL serialization enables VenusOCR~2.0 to recover the exact
cell alignment ($\mathrm{TEDS}=1.000$), while PaddleOCR-VL-1.6 hallucinates spurious spans and scores
$\mathrm{TEDS}=0.738$.

\clearpage

\end{document}